\PassOptionsToPackage{numbers, compress}{natbib}
\documentclass{article}

 \usepackage[main, final]{neurips_2026}

\usepackage[utf8]{inputenc} 
\usepackage[T1]{fontenc}    
\usepackage{amsmath}
\usepackage{hyperref}       
\usepackage{url}            
\usepackage{booktabs}       
\usepackage{amsfonts}       
\usepackage{nicefrac}       
\usepackage{microtype}      
\usepackage{xcolor}         

\usepackage{graphicx}
\usepackage{multirow} 
\usepackage{caption}
\usepackage{longtable}

\newcommand{\ra}[1]{\renewcommand{\arraystretch}{#1}} 
\usepackage{pifont}
\usepackage{makecell}
\usepackage[table]{xcolor}

\newcommand{\bs}[1]{\boldsymbol{#1}}
\newcommand{\mc}[1]{\mathcal{#1}}

\newcommand{\bbbr}{\mathbb{R}}

\newcommand{\egomore}{EgoMo3R}

\title{EgoMo3R: Joint Egocentric \\Motion and Scene Reconstruction}

\author{%
  Chaitanya Patel\textsuperscript{1} \quad
  Yusuke Kato\textsuperscript{1,2} \quad
  Rin Takahira\textsuperscript{3} \quad
  Kazuki Kozuka\textsuperscript{3} \\
  \textbf{Juan Carlos Niebles\textsuperscript{1}} \quad
  \textbf{Ehsan Adeli\textsuperscript{1}} \\[0.6em]
  \textsuperscript{1}Stanford University \quad
  \textsuperscript{2}Panasonic R\&D Company of America \\
  \textsuperscript{3}Panasonic Holdings Corporation \\[0.6em]
  \url{https://chaitanya100100.github.io/EgoMo3R/}
}

\begin{document}

\maketitle

\begin{abstract}
Understanding holistic 3D human motion and the surrounding environment from egocentric video is fundamental for applications in AR/VR and robotics. Although these tasks are inherently coupled, existing methods treat them separately. In particular, egocentric motion reconstruction methods typically assume access to metric, gravity-aligned camera trajectories from specialized inertial sensors. Recovering such trajectories from monocular egocentric video remains a significant challenge due to rapid head movements and motion blur. We present ~\egomore, a joint optimization framework that simultaneously reconstructs 3D human motion and scene geometry from casual egocentric video. It leverages the robustness of low-level visuo-geometric primitives, including monocular depth, optical flow, and point tracks, integrated with a diffusion-based human motion prior via first-order optimization. We propose an alternating optimization scheme that establishes a synergistic feedback loop: scene reconstruction provides physical grounding and trajectory conditioning for the motion prior, while the motion prior regularizes the camera trajectory and anchors the scene in a metric, gravity-aligned world frame. We evaluate ~\egomore~on the highly dynamic sequences of the EgoExo4D and Nymeria datasets and show that it outperforms specialized baselines in both 3D scene reconstruction and egocentric motion estimation.
\end{abstract}

\begin{figure}[t]
\centering
\includegraphics[width=\textwidth]{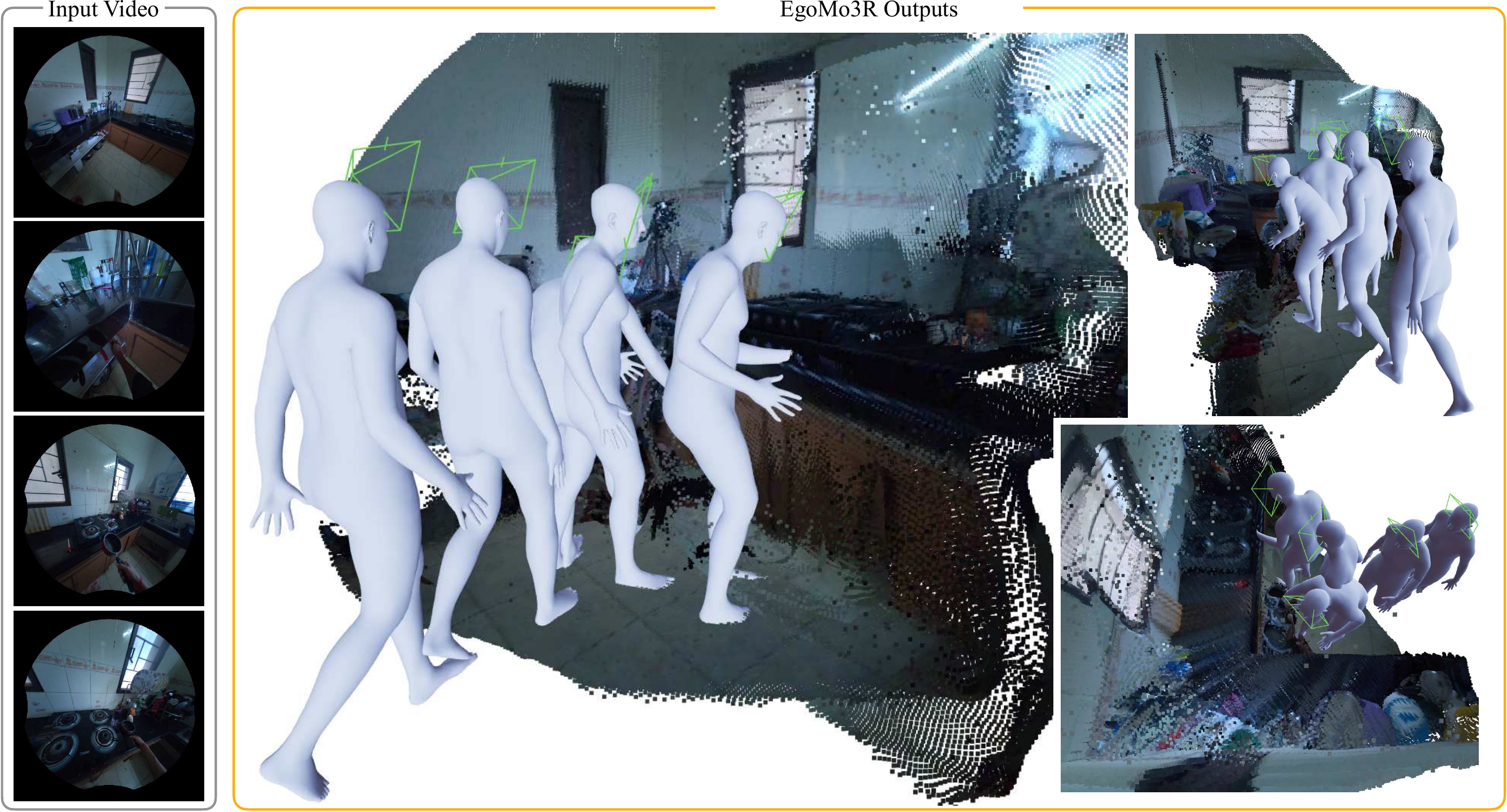}
\caption{Given an egocentric video, ~\egomore~optimizes for \textcolor{black!30!green}{camera poses}, 3D pointmaps, and \textcolor{white!50!blue}{human motion} in a common gravity-aligned metric reference frame. We also show two additional views of our scene and motion reconstruction on the side.}
\label{fig:teaser}
\vspace{-5mm}
\end{figure}

\section{Introduction}

A holistic 3D understanding of human motion and the surrounding environment from egocentric video is crucial for human-centered applications such as AR/VR, assistive technologies, and healthcare, while also offering a scalable pathway for robot learning. In the egocentric setting, 3D human motion and scene reconstruction are inherently coupled: recovering the user's full-body motion requires precise trajectory estimation of the egocentric camera that naturally follows the head motion. Traditional \textit{visual} SLAM methods~\cite{cadena2017past,davison2007monoslam,mur2015orb,droidslam,dpvo} typically treat camera trajectory estimation as a second-order optimization problem relying on multi-view correspondences to optimize for camera poses and sparse geometric representations. However, these methods often struggle on dynamic egocentric \textit{locomotion} videos with narrow fields of view and motion blur. This `inside-out' perspective results in sparse, low-quality correspondences leading to an underconstrained optimization, particularly during the rapid, rotation-heavy head movements common in daily activities~\cite{carlone2015initialization}. While recent data-driven feedforward reconstruction methods~\cite{vggt,cut3r} offer a promising alternative, they often struggle to generalize due to the egocentric domain gap. Consequently, existing egocentric motion reconstruction methods largely bypass these geometric challenges by assuming access to a perfect metric and gravity-aligned camera trajectories from a dedicated \textit{inertial} SLAM pipeline~\cite{egoallo,uniegomotion,avatarposer}. While joint modeling of humans and 3D scenes has shown promise in third-person settings~\cite{hsfm,human3r,rojas2025hamst3r}, these benefits do not trivially translate to the egocentric domain, where the human body remains largely invisible.

In this paper, we take the first step towards the joint modeling of egocentric motion and scene geometry with \egomore~(see Fig.~\ref{fig:teaser}). We demonstrate that \egomore~enables the synergistic reconstruction of both 3D human motion and the surrounding environment from a casually captured egocentric video. At its core, our approach formulates this task as a first-order joint optimization problem that recovers camera poses, scene geometry, and human motion in a single reference frame by leveraging data-driven low-level geometry and motion priors.

Our first insight is that while feedforward 3D reconstruction models often struggle with the egocentric domain gap, low-level visuo-geometric primitives, such as monocular depth, optical flow, and point tracks, exhibit significantly higher robustness. Although these cues can be individually noisy and mutually inconsistent, they provide complementary geometric constraints that, when aggregated within a global optimization framework, yield a coherent reconstruction. Specifically, our formulation utilizes monocular depth as a geometric base and point tracking for long-term temporal correspondence. To ensure convergence, we leverage optical flow to initialize the camera trajectory, effectively guiding our first-order optimization away from the poor local minima typical of rapid, rotation-heavy egocentric locomotion sequences.

Our second key insight is that egocentric human motion and scene reconstruction can be mutually reinforcing. Accurate 3D scene reconstruction with a precise camera trajectory directly informs head motion and offers explicit physical grounding for the body within the 3D environment. Conversely, 3D human motion serves as a unique prior for camera trajectory estimation, beyond simple constraints such as constant velocity or smoothness, common in traditional SLAM. Human motion priors also anchor the optimization in a canonical upright (metric, gravity-aligned, grounded) world frame, thereby improving the reconstruction on challenging egocentric videos.

However, unlike exocentric scenarios, jointly optimizing human motion with scene geometry is non-trivial in the egocentric domain, where the body remains largely out of view. To address this, we employ a motion diffusion generative model as a robust motion prior. In particular, we use a motion diffusion model conditioned on $\text{SE}(3)$-invariant egocentric trajectory and formulate an alternating optimization framework that iteratively refines the 3D scene and human motion. This setup creates a synergistic feedback loop: improved camera pose provides more accurate conditioning for the motion prior, which in turn improves physically grounded reconstruction. We evaluate \egomore~on EgoExo4D~\cite{egoexo4d} and Nymeria~\cite{nymeria} datasets and show that it outperforms strong baselines in both motion and scene reconstruction. Finally, we provide an extensive ablation of our optimization framework, including the technical design and integration of neural motion priors into our reconstruction pipeline.

In summary, we make the following contributions: (1) We present a joint first-order optimization framework that simultaneously recovers metric camera poses, scene geometry, and human motion from a single egocentric video. (2) We show that integrating complementary low-level visuo-geometric primitives and motion prior mitigates the challenging egocentric domain gap. (3) Our alternating optimization incorporates a diffusion-based motion prior to provide physical grounding and gravity alignment for both the human motion and the reconstructed scene. (4)~\egomore~outperforms strong specialized baselines in both 3D scene reconstruction and egocentric motion estimation.

\section{Related Works}

\textbf{Egocentric Motion Reconstruction:}
The front-facing nature of egocentric RGB cameras~\cite{projectaria} presents significant challenges for motion reconstruction due to limited body visibility. To compensate for this, many works use simulation-based physical motion priors~\cite{kinpoly,egopose,egoposebefore}, data-driven motion diffusion priors~\cite{uniegomotion,egoego,egoallo,avatarposer,guzov2025hmd2,bodiffusion}, or autoregressive motion priors~\cite{wang2025ego4o,hong2025egolm} learned from large-scale motion datasets. However, these priors critically depend on an accurate, metric, and gravity-aligned egocentric camera trajectory.
Several recent works~\cite{uniegomotion,egoallo,hamos,guzov2025hmd2} sidestep this challenge by assuming access to such trajectories from highly optimized inertial SLAM systems~\cite{projectaria} with multi-modal sensors, which are not always available in casual capture scenarios. Although classical monocular SLAM or VIO~\cite{droidslam,dpvo} methods can provide an $\text{SE}(3)$-invariant trajectory, these estimates are typically inaccurate and prone to significant jitter in egocentric setup. To remedy this, some methods~\cite{egoego,kinpoly,egopose,egoposebefore} attempt to learn camera pose estimation directly from optical flow features. We demonstrate that such flow-based learned pose estimation models struggle with egocentric locomotion in cluttered scenes, which subsequently degrades downstream motion reconstruction. In contrast,~\egomore~presents a robust optimization framework that simultaneously recovers camera poses, 3D scene geometry, and human motion from casual egocentric videos in a synergistic manner.

\textbf{Camera Pose Estimation \& 3D Reconstruction:}
Classic SfM~\cite{colmap, colmap2, glomap, romeinaday} and SLAM~\cite{cadena2017past, davison2007monoslam, mur2015orb} frameworks typically formulate camera pose and 3D scene estimation as a second-order bundle adjustment problem, relying on multi-view correspondences from locally invariant keypoint descriptors~\cite{sift, orb}. To improve robustness, various approaches integrate data-driven components ranging from learned sparse features~\cite{sarlin2020superglue, detone2018superpoint} to dense correspondence learning~\cite{dpvo, droidslam, tateno2017cnnslam}. Recent methods~\cite{zhao2022particlesfm, wang2024vggsfm, smith2025flowmap, megasam, vipe, shapeofmotion} further leverage robust priors, such as long-term point tracking~\cite{cotracker, doersch2023tapir, cotracker3}, monocular depth~\cite{moge, dav2}, video depth~\cite{video_depth_anything}, and semantic segmentation~\cite{sam, sam2}, to handle dynamic scenes and sparse views.
While several works have adapted similar strategies to the egocentric setting~\cite{egogaussian, egom2p, egomono4d}, they struggle on egocentric locomotion videos with rapid, rotation-heavy camera movement.
Although recent end-to-end feedforward models~\cite{dust3r, mast3r, zhang2024monst3r, vggt, cut3r} are promising due to their simplicity, their performance often degrades on challenging egocentric inputs.
Closing this gap through finetuning is difficult since existing egocentric datasets lack large-scale, dense, high-quality metric 3D groundtruth.
\egomore~addresses these challenges by combining robust visuo-geometric primitives with a diffusion human motion prior, ensuring accurate, physically grounded, and gravity-aligned reconstruction.

\textbf{Joint Human and Scene Reconstruction:}
Several works have leveraged the presence of humans within a scene to overcome the limitations of classical reconstruction approaches. For instance, ~\citet{ma2022humanextremeview} and ~\citet{xu2021widepersonreid} use 3D human pose estimation and person re-identification, respectively, to define virtual correspondences for geometry estimation in extreme views. Recent methods~\cite{pavlakos2022one,hsfm,rojas2025hamst3r,liu2025josh3r} perform joint 3D reconstruction of humans and scenes by optimizing global objectives on top of independent predictions. More recently, Human3R~\cite{human3r} builds upon CUT3R~\cite{cut3r} using visual prompting to efficiently train for simultaneous 3D human mesh and scene reconstruction in a feedforward manner. 
These approaches demonstrate that joint reconstruction is not only achievable but that the two tasks can be mutually beneficial~\cite{zhao2024synergistichumanscene,human3r}. However, such methods rely on explicit human-related visual cues within the image. In contrast, the user’s body remains largely invisible in egocentric viewpoints. In this setting, human-specific cues are not encoded in the raw pixels, but rather in the camera motion that generated them. ~\egomore~leverages these implicit motion cues for the synergistic reconstruction of humans and scenes in the egocentric domain. 
\section{Method}

\subsection{Problem Formulation}
Given an egocentric video sequence $\bs I_{1:N} \in \bbbr^{N \times H \times W \times 3}$ captured by a head-mounted camera, our goal is to jointly estimate camera trajectory, 3D scene geometry, and the underlying 3D human motion in a \textit{physically grounded} canonical world frame with known gravity direction and ground plane. We denote the camera poses as extrinsics $\bs C \in \bbbr^{N \times 4 \times 4}$ representing camera-to-world $SE(3)$ transformations. The 3D scene geometry is represented by a sequence of dense pointmaps~\cite{dust3r} $\bs P \in \bbbr^{N \times H \times W \times 3}$, where each element $\bs P_{n,i,j} \in \bbbr^3$ corresponds to the 3D coordinates of pixel $(i, j)$ at time $n$ in the local camera coordinate frame. The global 3D scene representation $\bs P^W$ in the world frame is obtained by transforming local pointmaps $\bs P$ using the estimated camera poses $\bs C$.
The 3D human motion $\bs M_{1:N}$ is defined by SMPL-X parameters~\cite{smplx} $M_n = (\bs{R}_n^r, \bs{t}_n^r, \theta_n)$ comprising root rotation ${R}_n^r \in \bbbr^3$, translation ${t}_n^r \in \mathbb{R}^3$ (at pelvis), and body pose angles $\theta_n \in \bbbr^{21 \times 3}$.

\subsection{Visual Geometry Priors}
\label{sec:vis_geo_priors}
While end-to-end egocentric reconstruction often struggles with egocentric domain shift, low-level visuo-geometric primitives offer a more robust basis for optimization. We formulate a first-order optimization framework that treats these primitives as soft constraints.
Rather than treating 3D points as free variables, we adopt per-frame relative depth $\bs{D}^r \in \mathbb{R}^{N \times H \times W}$, predicted by an off-the-shelf monocular depth estimator~\cite{moge2}, as our base geometry. 
While modern monocular estimators~\cite{moge, moge2, dav2} provide robust relative depth, their metric scale consistency often degrades in egocentric settings.
To address this, we optimize per-frame scale factors $\bs{S} \in \mathbb{R}^{N}$ to derive metric depth estimates $D_n = S_n D_n^r$. 
2D pixels in $\bs{I}$ can now be backprojected into 3D using depth $\bs{D}$ and known camera intrinsics $K \in \mathbb{R}^{3 \times 3}$ to obtain camera-frame pointmaps $\bs{P}$ as $P_{n,i,j} = K^{-1} [i, j, D_{n,i,j}]^T$.

Additionally, we use an off-the-shelf optical flow model~\cite{raft} to predict dense optical flow $\bs O \in \bbbr^{N \times H \times W \times 2}$, where $\bs O_{n,i,j} = (\delta u, \delta v)$ implies that pixels $\bs I_{n,i,j}$ and $\bs I_{n+1,i+\delta u, j+\delta v}$ correspond to the same 3D point. Formally, this induces the constraint $ C_{n+1}  P_{n+1,i+\delta u, j+\delta v} =  C_n  P_{n,i,j}$. While these constraints are theoretically sufficient to estimate $\bs C$, they only provide local supervision between consecutive frames. In egocentric settings, monocular depth and optical flow often degrade due to motion blur and narrow fields of view. Consequently, small errors compound over time, leading to significant trajectory drift and distorted 3D reconstructions. Traditional SLAM pipelines~\cite{davison2007monoslam,droidslam,vipe,megasam} typically employ a two-tiered system to correct such drift via loop closure. Instead, we leverage an off-the-shelf 2D point tracking model~\cite{cotracker3} to provide robust, long-range correspondences $\bs T$ to maintain reconstruction alignment~\cite{zhao2022particlesfm,smith2025flowmap,shapeofmotion,egomono4d}. We demonstrate in our experiments that relying on visuo-geometric priors already yields strong performance in camera pose estimation and 3D scene reconstruction. However, these purely visual constraints remain insufficient for recovering the underlying human motion, gravity direction, or floor grounding.

\subsection{Motion Prior}

The egocentric setup is unique in that the camera trajectory $\bs C$ directly follows the human head motion.
Prior works~\cite{egoego,egoallo,uniegomotion} have leveraged this to perform egocentric human motion reconstruction, often without access to paired datasets, achieving impressive results even when the body largely remains out-of-view. These methods typically formulate motion reconstruction as sampling from $p(\bs M \vert \bs C)$, where $\bs C$ is assumed to be a metric, gravity-aligned, and floor-grounded camera trajectory. This conditional distribution $p(\bs M \vert \bs C)$ is usually modeled via a conditional motion diffusion model $f_\phi(\bs M_t, \bs C)$ with learnable parameters $\phi$, where $\bs M_t$ denotes the noisy motion at diffusion timestep $t$.

However, incorporating a strong diffusion-based motion prior into our joint optimization framework is non-trivial for several reasons. First, unlike prior works, we do not have \textit{a priori} access to a gravity- and floor-aligned camera trajectory $\bs C$.
Furthermore, training unconditional diffusion models remains inherently difficult for complex distributions such as images~\cite{returnofuncond,hu2023selfguided} and human motion~\cite{raab2023modi,mootuncondmotion}. The lack of direct body observations makes it difficult to guide the denoising process to adhere to a suboptimal, unaligned trajectory $\bs C$ that is itself undergoing optimization (See the Appendix for more discussion).

To remedy this, we instead model the distribution $p(\bs M \vert \mc{A}(\bs C))$, where $\mc{A}(\bs C)$ is an $\text{SE}(3)$-invariant canonicalized camera trajectory defined as $\mc{A}(\bs C)_i =  C_0^{-1}  C_i$. We train a conditional diffusion model $f_\phi(\bs M_t, \mc{A}(\bs C))$ to generate physically grounded motion $\bs M$ conditioned on this relative trajectory. This design is motivated by the observation that the $\text{SE}(3)$-invariant trajectory remains highly informative of the underlying body motion. Crucially, it allows for the stable integration of the motion prior during the early stages of optimization when $\bs C$ is not yet optimal or canonically aligned. It not only provides unique prior on the camera trajectory $\bs C$ but also improves the overall 3D reconstruction by implicitly enforcing gravity direction and ground-plane priors.

\subsection{Optimization}

\begin{figure}[t]
\centering
\includegraphics[width=0.99\textwidth]{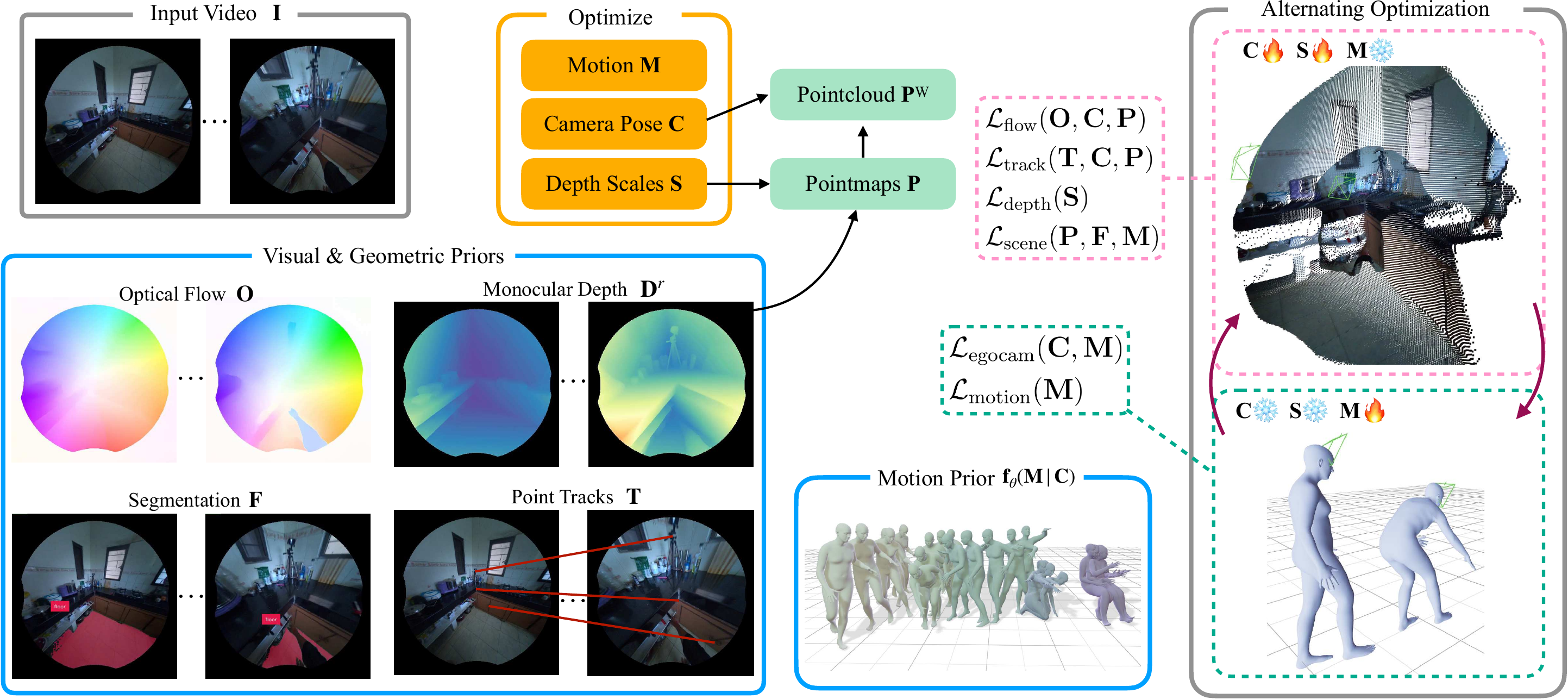}
\caption{ Given an input video $\bs I$, we leverage a monocular depth estimator to obtain relative depth $\bs D^r$ as a geometric base, optimizing only the per-frame scales $\bs S$ to derive metric depth $\bs D$. Our framework jointly optimizes for camera poses $\bs C$, which transform metric depth into a global world-frame pointmaps $\bs P^W$, and 3D human motion $\bs M$. We employ an alternating optimization strategy: (1) \textbf{Scene Refinement} updates $\bs C$ and $\bs S$ using optical flow, point tracks, and a scene-alignment loss to anchor the reconstruction in a canonical frame provided by motion; (2) \textbf{Motion Alignment} refines $\bs M$ by leveraging a diffusion motion prior conditioned on $\bs C$. This synergistic loop prevents optimization collapse and ensures physically grounded joint 3D scene and motion reconstruction.}
\label{fig:overview}
\vspace{-5mm}
\end{figure}

Given visual primitives and motion prior, we formulate the following optimization problem:
\begin{equation}
(\bs C^*, \bs M^*, \bs S^*) =
\underset{\bs C, \bs M, \bs S}{\operatorname{argmin}}\;\;
\mc{L}_\text{flow} +
\mc{L}_\text{track} +
\mc{L}_\text{depth} +
\mc{L}_\text{scene} +
\mc{L}_\text{egocam} +
\mc{L}_\text{motion}
\label{eq:basic_optim}
\end{equation}

The term $\mc{L}_\text{flow} = \sum_{n,i,j} \| O_{n,i,j} - \Pi_K( C^{-1}_{n+1} C_n  P_{n,i,j}) \|_2^2$ minimizes the reprojection error between the observed optical flow and the flow induced by the estimated camera motion, where $\Pi_K$ denotes the camera projection using intrinsics $K$. Similarly, $\mc{L}_\text{track}$ enforces long-range geometric consistency across non-consecutive frames using the point tracks $\bs T$. We use Median absolute deviation filtering on $\mc{L}_\text{flow}$ and $\mc{L}_\text{track}$ to mitigate the influence of bad correspondences from dynamic objects. To regularize the depth estimation, we define $\mc{L}_\text{depth} = \sum_{n}\|S_n - S_n^m\|_2^2$, which constrains the optimized per-frame scales $\bs S$ to remain near the initial metric scales $\bs S^m$ provided by an off-the-shelf estimator~\cite{moge2}. While $\bs S^m$ may suffer from monocular ambiguity in egocentric setting, this term prevents the optimization from converging to degenerate solutions in the presence of noisy or no correspondences.

The term $\mc{L}_\text{scene} = \sum_{n,i,j} F_{n, i, j} \cdot d(\bs M, P^W_{n, i, j})$ ensures the canonical world frame alignment between points $\bs P^W$ and motion $\bs M$, where $\bs F$ is a binary ground segmentation mask~\cite{groundingdino,sam2} and $d(\bs M, a)$ computes the point-to-plane distance between a 3D point $a$ and the ground plane implicitly defined by the human motion $\bs M$.
The coupling between the camera and the human motion is enforced via $\mc{L}_\text{egocam} = \| \bs C - \mathcal{H}(\bs M) \|_2^2$, which aligns the camera trajectory $\bs C$ with the head trajectory $\mathcal{H}(\bs M)$ derived from the SMPL-X parameters. Finally, we incorporate the generative motion prior via a denoising score-matching loss $\mc{L}_\text{motion} = \mathbb{E}_{t\sim\mathcal{U}[0, 1]} \left[\| f_\phi(\bs M_t, \mc{A}(\bs C)) - \bs M\|_2^2\right]$. It drives the motion $\bs M$ toward the manifold of physically plausible human movements conditioned on $\mc{A}(\bs C)$.

\subsection{Alternating Optimization}

We observe that directly minimizing Eq.~\ref{eq:basic_optim} via gradient descent often yields suboptimal results and poor alignment between the camera trajectory $\bs C$ and human motion $\bs M$. This instability stems primarily from two factors. First, tracking failures can introduce significant outlier correspondences that bias the optimization. While we employ Median Absolute Deviation (MAD) filtering -- analogous to RANSAC~\cite{ransac} in traditional SLAM -- a poor initialization can lead to the erroneous rejection of valid correspondences, causing the optimization to converge to a local minimum. To mitigate this, we initialize the camera trajectory by optimizing $\bs C$ solely with the $\mc{L}_\text{flow}$ objective. As discussed in Sec.~\ref{sec:vis_geo_priors}, $\mc{L}_\text{flow}$ is a 2D reformulation of the 3D constraint $ C_{n+1} P_{n+1,i+\delta u, j+\delta v} = C_n P_{n,i,j}$, which has a closed-form solution using Procrustes analysis~\cite{procrustes}. This warm-start enables more accurate outlier detection and stabilizes subsequent optimization steps. Second, the simultaneous optimization of $\bs C$ and $\bs M$ introduces noisy gradients from $\mc{L}_\text{motion}$, as our diffusion prior $f_\phi(\bs M_t, \mc{A}(\bs C))$ is itself a function of the evolving camera trajectory $\bs C$~\cite{li2026noperoomgs}. To resolve this, we propose an alternating optimization scheme that iterates between two distinct modes:

\textbf{1. Camera and Scene Refinement:} We optimize the camera poses $\bs C$ and depth scales $\bs S$ using $\mc{L}_\text{flow}$, $\mc{L}_\text{track}$, $\mc{L}_\text{depth}$, and $\mc{L}_\text{scene}$, while keeping the human motion $\bs M$ fixed. This stage refines the 3D scene reconstruction while leveraging the gravity and ground-plane priors from fixed motion $\bs M$.

\textbf{2. Motion and Prior Alignment:} We optimize only the motion parameters $\bs M$ using $\mc{L}_\text{motion}$ and $\mc{L}_\text{egocam}$, while keeping $(\bs C, \bs S)$ fixed. This stage enforces the generative motion prior and eliminates physical implausibilities in the camera trajectory, such as jitters or sudden jumps. It also defines the canonical frame to improve 3D reconstruction. This mode can be interpreted as sampling from the posterior $p(\bs M \vert \mc{A}(\bs C))$ with guidance~\cite{diffusionbeatsgans} provided by the camera-head alignment loss $\mc{L}_\text{egocam}$.

We demonstrate empirically that this alternating strategy significantly improves joint reconstruction results compared to direct optimization. Please refer to the Appendix for more details.

\section{Experiments}

\subsection{Dataset}
We evaluate our approach on {EgoExo4D}~\cite{egoexo4d} and {Nymeria}~\cite{nymeria} datasets. Both datasets provide multimodal egocentric data captured via Project Aria smartglasses~\cite{projectaria}. The Aria MPS kit provides high-quality groundtruth camera poses obtained through an extensive offline VIO and SLAM system that fuses data from multiple sensors~\cite{projectaria}.
All frames are undistorted to a linear pinhole camera following the official Project Aria guidelines~\cite{projectaria}, at $512\times512$ resolution with a focal length of 150\,px and without cropping. All methods, including \egomore{}, receive these identical rectified frames.
To evaluate 3D reconstruction, we filter the MPS SLAM point clouds using 2D observation metadata to retain only high-confidence points. 
For 3D human motion, Nymeria provides groundtruth motion annotations, while for EgoExo4D, we use the 3D human motion fits provided by UniEgoMotion~\cite{uniegomotion} as groundtruth. For our testing, we sample video clips uniformly at random from both datasets. To further assess performance in challenging locomotion scenarios, we additionally include segments where groundtruth camera poses exhibit high movement.
We report metrics on this combined test set in the main paper, while separate metrics on the uniform and locomotion-heavy subsets are provided in the Appendix.

\subsection{3D Reconstruction \& Camera Pose Estimation Results}

\begin{table*}[t]\centering\footnotesize
\setlength{\tabcolsep}{6.0pt}
\ra{1.1} 
\caption{\textbf{3D Reconstruction}: Comparison across metric and scale-invariant 3D reconstruction. Lower values ($\downarrow$) indicate better performance. Our method outperforms all baselines by a significant margin.}
\label{tab:3d_recon}
\begin{tabular}{llcccccccc} 
\toprule
& & \multicolumn{4}{c}{Metric Scale} & \multicolumn{4}{c}{Scale Invariant} \\
\cmidrule(lr){3-6} \cmidrule(lr){7-10}
& & \multicolumn{2}{c}{Acc} & \multicolumn{2}{c}{Comp} & \multicolumn{2}{c}{Acc} & \multicolumn{2}{c}{Comp} \\
& Method & Mean & Median & Mean & Median & Mean & Median & Mean & Median \\
\midrule
\multirow{5}{*}{\rotatebox[origin=c]{90}{\textbf{EgoExo4D}}} 
& VGGT~\cite{vggt} & $-$ & $-$ & $-$ & $-$ & $0.470$ & $0.418$ & $0.440$ & $0.226$ \\
& CUT3R~\cite{cut3r} & $\underline{0.433}$ & $\underline{0.397}$ & $\underline{0.665}$ & $\underline{0.506}$ & $\underline{0.410}$ & $\underline{0.350}$ & $\underline{0.341}$ & ${0.226}$ \\
& ViPE~\cite{vipe} & $0.905$ & $0.657$ & $0.819$ & $0.573$ & $0.425$ & $0.366$ & $0.352$ & $\underline{0.194}$ \\
& EgoMono4D~\cite{egomono4d} & $0.619$ & $0.544$ & $0.707$ & $0.546$ & $0.469$ & $0.423$ & $0.593$ & $0.413$ \\
& Ours & $\mathbf{0.357}$ & $\mathbf{0.287}$ & $\mathbf{0.187}$ & $\mathbf{0.124}$ & $\mathbf{0.309}$ & $\mathbf{0.245}$ & $\mathbf{0.158}$ & $\mathbf{0.098}$ \\
\midrule 
\multirow{5}{*}{\rotatebox[origin=c]{90}{\textbf{Nymeria}}} 
& VGGT~\cite{vggt} & $-$ & $-$ & $-$ & $-$ & $\underline{0.337}$ & ${0.290}$ & $0.309$ & ${0.151}$ \\
& CUT3R~\cite{cut3r} & $\underline{0.346}$ & $\underline{0.298}$ & $\underline{0.362}$ & $\underline{0.229}$ & $0.338$ & $\underline{0.286}$ & $0.293$ & $0.180$ \\
& ViPE~\cite{vipe} & $1.742$ & $1.520$ & $0.762$ & $0.566$ & $0.345$ & $0.298$ & $\underline{0.271}$ & $\underline{0.141}$ \\
& EgoMono4D~\cite{egomono4d} & $0.489$ & $0.417$ & $0.556$ & $0.390$ & $0.395$ & $0.355$ & $0.544$ & $0.375$ \\
& Ours & $\mathbf{0.247}$ & $\mathbf{0.189}$ & $\mathbf{0.131}$ & $\mathbf{0.073}$ & $\mathbf{0.223}$ & $\mathbf{0.171}$ & $\mathbf{0.118}$ & $\mathbf{0.065}$ \\
\bottomrule
\end{tabular}
\vspace{-4mm}
\end{table*}

\begin{figure}[t]
\centering
\includegraphics[width=\textwidth]{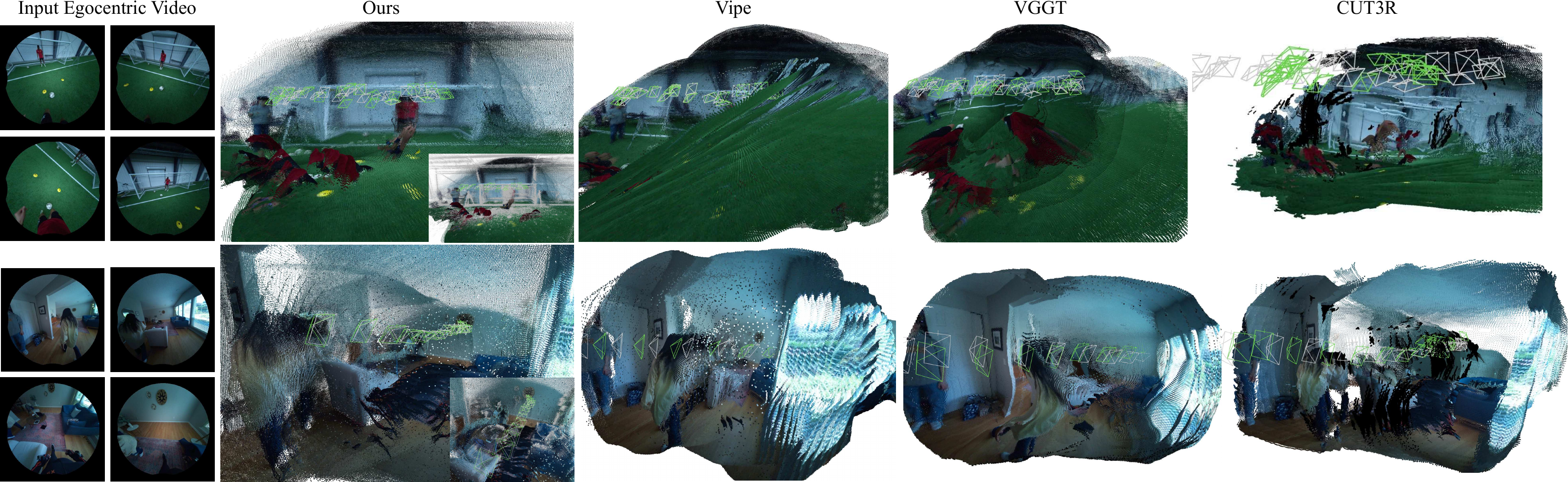}
\caption{\textbf{3D Reconstruction}: We visualize the predicted pointmaps and camera trajectories (green) against the groundtruth trajectories (grey). The inset shows our reconstructed scene overlaid with sparse groundtruth points (grey), demonstrating the metric accuracy and alignment of~\egomore. Please refer to our supplementary video for a better qualitative comparison.}
\label{fig:recon_vis}
\vspace{-4mm}
\end{figure}

\begin{table*}[t]\centering\footnotesize
\setlength{\tabcolsep}{10.0pt}
\ra{1.1} 
\caption{\textbf{Camera Pose Estimation}: Comparison across metric and scale-invariant camera pose estimation. Lower values ($\downarrow$) indicate better performance. 
}
\label{tab:camera_pose}
\begin{tabular}{llcccccc} 
\toprule
& & \multicolumn{3}{c}{Metric Scale} & \multicolumn{3}{c}{Scale Invariant} \\
\cmidrule(lr){3-5} \cmidrule(lr){6-8}
& Method & ATE & RPE Rot & RPE Trans & ATE & RPE Rot & RPE Trans \\
\midrule
\multirow{6}{*}{\rotatebox[origin=c]{90}{\textbf{EgoExo4D}}} 
& EgoEgo~\cite{egoego} & $0.421$ & $5.023$ & $\underline{0.089}$ & $0.275$ & $5.023$ & $0.067$ \\
& VGGT~\cite{vggt}    & - & - & - & $0.251$ & $3.744$ & $0.078$ \\
& CUT3R~\cite{cut3r} & $\underline{0.415}$ & $4.101$ & $0.118$ & $0.294$ & $4.101$ & $0.076$ \\
& ViPE~\cite{vipe}    & $0.426$ & $\mathbf{2.568}$ & $0.137$ & $\mathbf{0.112}$ & $\mathbf{2.568}$ & $\mathbf{0.032}$ \\
& EgoMono4D~\cite{egomono4d} & $1.351$ & $4.445$ & $0.347$ & $0.345$ & $4.445$ & $0.077$ \\
& Ours                & $\mathbf{0.226}$ & $\underline{2.640}$ & $\mathbf{0.065}$ & $\underline{0.161}$ & $\underline{2.640}$ & $\underline{0.047}$ \\
\midrule
\multirow{6}{*}{\rotatebox[origin=c]{90}{\textbf{Nymeria}}} 
& EgoEgo~\cite{egoego} & $0.629$ & $3.129$ & $0.096$ & $0.390$ & $3.129$ & $0.087$ \\
& VGGT~\cite{vggt}    & - & - & - & $0.275$ & $2.956$ & $0.107$ \\
& CUT3R~\cite{cut3r} & $0.471$ & $2.801$ & $0.119$ & $0.341$ & $2.801$ & $0.093$ \\
& ViPE~\cite{vipe}    & $\mathbf{0.276}$ & $\underline{2.168}$ & $\underline{0.085}$ & $\mathbf{0.122}$ & $\underline{2.168}$ & $\mathbf{0.039}$ \\
& EgoMono4D~\cite{egomono4d} & $1.318$ & $3.881$ & $0.322$ & $0.435$ & $3.881$ & $0.115$ \\
& Ours                & $\underline{0.325}$ & $\mathbf{2.142}$ & $\mathbf{0.060}$ & $\underline{0.239}$ & $\mathbf{2.142}$ & $\underline{0.065}$ \\
\bottomrule
\end{tabular}
\vspace{-3mm}
\end{table*}

\noindent\textbf{Baselines.}
\label{subsubsec:recon_baselines}
We compare \egomore's 3D reconstruction and camera pose estimation against several competitive baselines. 
{VGGT}~\cite{vggt} and {CUT3R}~\cite{cut3r} represent purely data-driven, general feedforward reconstruction methods trained on large-scale 3D data to directly regress camera poses and pointmaps. EgoMono4D~\cite{egomono4d} is a similar feedforward model trained on egocentric datasets.
{ViPE}~\cite{vipe} (similar to MegaSaM~\cite{megasam}) is a modern bundle-adjustment-based SLAM system with loop closure that optimizes 3D points as free variables while leveraging robust video depth, point tracks, and segmentation priors. 
{EgoEgo}~\cite{egoego} is the only baseline in the egocentric motion estimation literature that does not assume access to a canonical trajectory. Instead, it employs a dedicated camera pose estimation module on top of an off-the-shelf SLAM system~\cite{droidslam} and optical flow features. 
We train it on EgoExo4D training split and evaluate on both EgoExo4D and Nymeria to test its generalization capabilities. Notably, all other methods, including~\egomore, are evaluated without dataset-specific training.

\noindent\textbf{Metrics.}
For 3D reconstruction, we report standard {Accuracy} and {Completion} distances where lower values indicate superior performance.
To evaluate camera pose estimation, we report {Absolute Trajectory Error (ATE)}, which assesses the global accuracy after alignment with the ground truth. We also report {Relative Pose Error (RPE)} for both rotation and translation to evaluate local consistency. Since VGGT outputs are scale-agnostic while other methods (including ours) operate in metric scale, we report all metrics following both scale-preserving and scale-invariant alignment.

\noindent\textbf{Results.}
Tables~\ref{tab:3d_recon} and \ref{tab:camera_pose} present the results for 3D reconstruction and camera pose estimation, respectively. \egomore~ outperforms all baselines across all 3D reconstruction metrics by a significant margin. In camera pose estimation, our method performs on par with ViPE and outperforms all other baselines. Feedforward models such as VGGT and CUT3R struggle with the egocentric domain gap, where highly dynamic camera movements and significant motion blur lead to substantial errors in both reconstruction and pose estimation. EgoMono4D also performs poorly due to its reliance on only optical flow for correspondences. ViPE leverages low-level visuo-geometric priors including a video depth model, and uses classic loop closure to maintain accurate camera tracking, especially on texture-sparse Nymeria sequences. However, it produces overly smooth geometry, resulting in significantly worse 3D reconstruction metrics compared to \egomore. 
Finally, EgoEgo’s head-pose module performs poorly on our evaluation set. Its flow-based parameterization struggles with challenging head movements, particularly on the Nymeria sequences for which it was not trained. Overall, \egomore~demonstrates strong performance and robustness, achieving state-of-the-art results in both camera pose accuracy and 3D scene fidelity.

\subsection{Human Motion Reconstruction Results}

\begin{figure}[t]
\centering
\includegraphics[width=\textwidth]{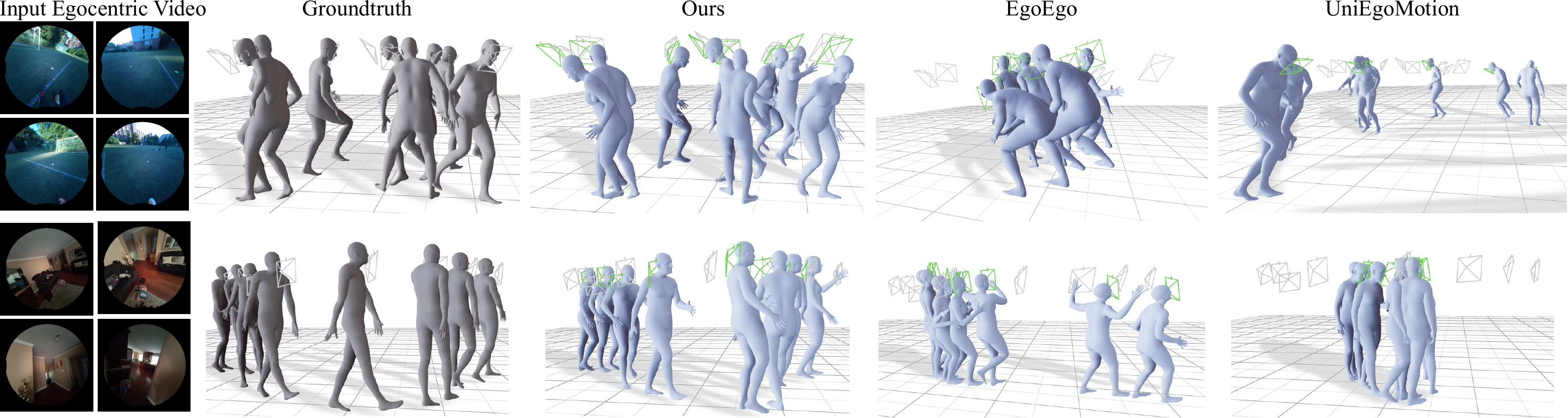}
\caption{\textbf{Qualitative comparison of 3D human motion reconstruction.} ~\egomore~effectively reconstructs physically grounded human motion that globally aligns well with the groundtruth. In contrast, errors in EgoEgo’s head trajectory estimation lead to significant global pose errors. While UniEgoMotion captures plausible local body poses, it similarly lacks the precise head tracking required for global accuracy. Please refer to our suppl. video for a better qualitative comparison.}
\label{fig:motion_vis}
\vspace{-2mm}
\end{figure}

\begin{table*}[t]\centering\footnotesize
\setlength{\tabcolsep}{8.0pt}
\ra{1.1}
\newcommand{\orc}[1]{\textcolor{gray}{#1}}
\caption{\textbf{Human Motion Reconstruction:} Lower values ($\downarrow$) indicate better performance. Our method significantly outperforms baselines on head tracking and global motion metrics, while baselines retain local pose accuracy at the cost of significantly worse alignment between the motion and the scene.}
\label{tab:human_motion}
\vspace{-2mm}
\begin{tabular}{llccccc}
\toprule
& Method & Head Rot & Head Trans & MPJPE & MPJPE-PA & FC \\
\midrule
\multirow{9}{*}{\rotatebox[origin=c]{90}{\textbf{\scriptsize EgoExo4D}}}
& EgoEgo~\cite{egoego} & $0.574$ & $0.449$ & $0.441$ & $0.143$ & ${0.046}$ \\
& CUT3R + EgoAllo~\cite{egoallo} & $0.278$ & $0.431$ & $0.427$ & $0.167$ & $\mathbf{0.035}$ \\
& CUT3R + UEM~\cite{uniegomotion} & $0.526$ & $0.456$ & $0.407$ & $0.151$ & $0.039$ \\
& ViPE + EgoAllo~\cite{egoallo} & $0.203$ & $0.411$ & $0.348$ & $0.155$ & $0.092$ \\
& ViPE + UEM~\cite{uniegomotion} & $0.531$ & $0.396$ & $0.337$ & $0.145$ & $0.088$ \\
& UEM-Img-only~\cite{uniegomotion} & $0.519$ & $0.334$ & $0.309$ & $\mathbf{0.073}$ & $0.037$ \\
& Ours & $\mathbf{0.188}$ & $\mathbf{0.219}$ & $\mathbf{0.271}$ & $0.127$ & $0.053$ \\
& \orc{GT Traj.\ + EgoAllo} & \orc{$0.119$} & \orc{$0.051$} & \orc{$0.174$} & \orc{$0.102$} & \orc{$0.033$} \\
& \orc{GT Traj.\ + UEM} & \orc{$0.203$} & \orc{$0.042$} & \orc{$0.127$} & \orc{$0.067$} & \orc{$0.029$} \\
\midrule
\multirow{9}{*}{\rotatebox[origin=c]{90}{\textbf{\scriptsize Nymeria}}}
& EgoEgo~\cite{egoego} & $0.538$ & $0.684$ & $0.600$ & $0.160$ & ${0.046}$ \\
& CUT3R + EgoAllo~\cite{egoallo} & $0.232$ & $0.488$ & $0.437$ & $0.178$ & $\mathbf{0.033}$ \\
& CUT3R + UEM~\cite{uniegomotion} & $0.448$ & $0.503$ & $0.442$ & $0.156$ & $0.035$ \\
& ViPE + EgoAllo~\cite{egoallo} & $0.222$ & $0.344$ & $0.341$ & $0.158$ & $0.084$ \\
& ViPE + UEM~\cite{uniegomotion} & $0.368$ & $0.365$ & $0.348$ & $0.157$ & $0.071$ \\
& UEM-Img-only~\cite{uniegomotion} & $0.806$ & $0.769$ & $0.696$ & $\mathbf{0.116}$ & $0.038$ \\
& Ours & $\mathbf{0.180}$ & $\mathbf{0.318}$ & $\mathbf{0.319}$ & $0.133$ & $0.036$ \\
& \orc{GT Traj.\ + EgoAllo} & \orc{$0.115$} & \orc{$0.070$} & \orc{$0.149$} & \orc{$0.108$} & \orc{$0.031$} \\
& \orc{GT Traj.\ + UEM} & \orc{$0.170$} & \orc{$0.068$} & \orc{$0.140$} & \orc{$0.103$} & \orc{$0.030$} \\
\bottomrule
\end{tabular}
\end{table*}

\begin{table*}[t]\centering\footnotesize
\setlength{\tabcolsep}{4.0pt}
\ra{1.1} 
\caption{\textbf{Ablation Study}: Metrics are reported on EgoExo4D locomotion subset with metric scale alignment. Lower values ($\downarrow$) are better. Rows 1--4 optimize only $(\bs C, \bs S)$. Rows 5--6 add the motion prior, optimized directly or in an alternating manner (ours, gray). Row 7 replaces the motion prior with a camera-smoothness prior.}
\label{tab:ablation_optim}
\vspace{-2mm}
\begin{tabular}{lccccccc} 
\toprule
 & \multirow{2}{*}{ATE} & \multirow{2}{*}{RPE Trans} & \multicolumn{2}{c}{Acc} & \multicolumn{2}{c}{Comp} & \multirow{2}{*}{MPJPE} \\
\cmidrule(lr){4-5} \cmidrule(lr){6-7}
 & & & Mean & Median & Mean & Median & \\
\midrule
$\;\;\mc{L}_\text{flow}$ & $0.604$ & $0.222$ & $0.426$ & $0.352$ & $0.207$ & $0.122$ & -- \\
$+\,\mc{L}_\text{track}$ & $0.361$ & $0.111$ & $0.418$ & $0.342$ & $0.213$ & $0.139$ & -- \\
$+\,\mc{L}_\text{depth}$ & $0.353$ & $0.103$ & $0.414$ & $0.337$ & $0.210$ & $0.135$ & -- \\
$+\,\mc{L}_\text{scene}$ & $0.371$ & $0.120$ & $0.398$ & $0.319$ & $0.211$ & $0.131$ & -- \\
$+\,\mc{L}_\text{motion} + \mc{L}_\text{egocam}$ {\scriptsize(direct)} & $0.350$ & $0.096$ & $0.403$ & $0.328$ & $0.214$ & $0.139$ & $0.456$ \\
\rowcolor{gray!20}
$+\,${\scriptsize Alternating Optim. (ours)} & $0.318$ & $0.095$ & $0.396$ & $0.320$ & $0.209$ & $0.137$ & $0.320$ \\
\midrule
Row 4 $+\,\mc{L}_\text{smooth}$ & $0.335$ & $0.074$ & $0.429$ & $0.351$ & $0.223$ & $0.142$ & -- \\
\bottomrule
\end{tabular}
\vspace{-4mm}
\end{table*}


\begin{table*}[t]\centering\footnotesize
\setlength{\tabcolsep}{4.0pt}
\ra{1.1} 
\caption{\textbf{Progression of Alternating Optimization}: Metrics are reported on EgoExo4D locomotion subset with metric scale alignment. Lower values ($\downarrow$) are better. Compared to direct optimization, alternation prevents the motion $\bs M$ from collapsing to a suboptimal solution while also improving camera pose and scene reconstruction. We report results after two rounds throughout the paper (gray rows). A third round further improves pose, accuracy, and MPJPE.}
\label{tab:alternating_ablation}
\begin{tabular}{lccccccc} 
\toprule
 & & & \multicolumn{2}{c}{Acc} & \multicolumn{2}{c}{Comp} & \\
\cmidrule(lr){4-5} \cmidrule(lr){6-7}
 & ATE & RPE Trans & Mean & Median & Mean & Median & MPJPE \\
\midrule
Direct optimization & $0.350$ & $0.096$ & $0.403$ & $0.328$ & $0.214$ & $0.139$ & $0.456$ \\
\midrule
\multicolumn{8}{l}{\textit{Alternating optimization}} \\
Initialization & $0.604$ & $0.222$ & $0.426$ & $0.352$ & $0.207$ & $0.122$ & -- \\
R1. Optimize $(\bs C, \bs S)$ & $0.353$ & $0.102$ & $0.410$ & $0.333$ & $0.212$ & $0.137$ & -- \\
R1. Optimize $(\bs M)$ & -- & -- & -- & -- & -- & -- & $0.347$ \\
\rowcolor{gray!20}R2. Optimize $(\bs C, \bs S)$ & $0.318$ & $0.095$ & $0.396$ & $0.320$ & $0.209$ & $0.137$ & -- \\
\rowcolor{gray!20}R2. Optimize $(\bs M)$ & -- & -- & -- & -- & -- & -- & $0.320$ \\
R3. Optimize $(\bs C, \bs S)$ & $0.300$ & $0.085$ & $0.391$ & $0.314$ & $0.209$ & $0.137$ & -- \\
R3. Optimize $(\bs M)$ & -- & -- & -- & -- & -- & -- & $0.315$ \\
\bottomrule
\end{tabular}
\end{table*}

\noindent\textbf{Baselines.}
We evaluate human motion reconstruction of~\egomore~against baseline methods that don't rely on a metric gravity-aligned camera trajectory.
For EgoEgo~\cite{egoego}, its head pose estimator is trained on optical flow features from the EgoExo4D dataset (see Sec.~\ref{subsubsec:recon_baselines}), while the gravity direction predictor is trained on AMASS~\cite{amass}.
For UniEgoMotion~\cite{uniegomotion} and EgoAllo~\cite{egoallo}, we obtain a metric camera trajectory from CUT3R or ViPE and estimate the floor plane using their floor-segmented~\cite{sam2} pointmaps.
Since floor estimation depends on reconstruction quality, CUT3R offers a balance of pose and geometry accuracy, whereas ViPE provides more accurate poses but weaker geometry.
As an oracle upper bound, we also condition both methods on groundtruth trajectories.
Given UniEgoMotion's support for flexible inputs, we also train its variant on EgoExo4D that directly reconstructs human motion from input video without requiring trajectory input.
Both EgoEgo and UniEgoMotion baselines are trained on the EgoExo4D training set and evaluated on both EgoExo4D and Nymeria sequences to analyze their generalization capacity.


\noindent\textbf{Metrics.}
We evaluate methods on their ability to track head motion using head rotation and head translation error~\cite{uniegomotion,egoego}.
MPJPE calculates mean per-joint position error (in meters) after globally aligning predicted body joints with groundtruth body joints.
MPJPE-PA is calculated after separately aligning each frame's motion and only measures the local human pose accuracy regardless of its global alignment.
FC (Foot Contact) ~\cite{uniegomotion} quantifies the plausibility of foot-ground contact.

\noindent\textbf{Results.}
Tab.~\ref{tab:human_motion} presents the human motion reconstruction results.~\egomore~outperforms the baselines on both head tracking and global pose metrics.
UEM-Img-only struggles to generate accurate global motion because it attempts to jointly estimate camera trajectories and body poses within a single model without explicit geometric constraints.
As discussed in Sec.~\ref{subsubsec:recon_baselines}, EgoEgo's head pose estimation pipeline struggles on challenging in-the-wild egocentric sequences.
Similarly, EgoAllo and UEM depend on imperfect camera trajectories provided by CUT3R/ViPE, causing errors to propagate into downstream motion estimation.
With groundtruth trajectories, both EgoAllo and UEM improve substantially, indicating that a major portion of the motion error is induced by trajectory error.
In contrast,~\egomore's joint optimization establishes a feedback loop that iteratively refines both camera poses and human motion. The resulting more accurate head trajectory translates into lower global motion error, even under loose $\text{SE}(3)$-invariant conditioning.
Interestingly, the baselines achieve competitive local pose accuracy and foot contact scores because their motion diffusion models strictly follow the potentially inaccurate input \textit{canonical} trajectory to generate plausible human motion.


\subsection{Ablation}
\label{sec:ablation}
Tab.~\ref{tab:ablation_optim} isolates each loss term and the role of the motion prior. Without the motion prior, there is no motion stage to alternate with, so rows 1--4 are the scene-only variants of our method. Optimizing $\mc{L}_\text{flow}$ alone leads to drift, which $\mc{L}_\text{track}$ largely corrects through long-range correspondences (ATE $0.604\!\to\!0.361$). $\mc{L}_\text{depth}$ further regularizes the depth scales $\bs S$. $\mc{L}_\text{scene}$ enforces floor coplanarity, which improves reconstruction accuracy but raises ATE. Adding the motion prior with direct joint optimization (row 5) improves the camera trajectory but often causes the motion to collapse (MPJPE $0.456$). Alternating between the $(\bs C, \bs S)$ and $\bs M$ stages stabilizes the optimization and gives the lowest ATE, Acc Mean, and MPJPE (row 6). Replacing the motion prior with a camera-smoothness prior $\mc{L}_\text{smooth}$ (row 7) suppresses camera jumps and gives the lowest RPE Trans, but it is worse than our full model on all other metrics and produces no body motion.

Tab.~\ref{tab:alternating_ablation} shows the steady progression of the alternating optimization and that the coupling works in both directions. \emph{Motion$\to$scene:} the first motion stage lowers ATE from $0.353$ after the R1 scene stage to $0.318$ after R2, as the prior anchors the trajectory in a gravity-aligned, grounded frame. \emph{Scene$\to$motion:} MPJPE decreases monotonically as $(\bs C, \bs S)$ is refined across rounds ($0.347\!\to\!0.320\!\to\!0.315$), as a better trajectory gives better conditioning to the motion prior. A third round further improves pose, accuracy, and MPJPE.
\section{Conclusion \& Future Work}

We present~\egomore, an optimization framework that jointly reconstructs egocentric human motion, camera trajectory, and metric, gravity-aligned scene geometry from casually captured monocular egocentric video. By combining low-level visuo-geometric cues with a diffusion-based motion prior conditioned on an $\text{SE}(3)$-invariant representation of the camera trajectory, our approach exploits the inherent coupling between camera trajectory, scene geometry, and body pose in the egocentric setting.~\egomore~outperforms specialized baselines on the EgoExo4D and Nymeria datasets, and our ablations show that a learned motion prior outperforms a simple smoothness penalty. This suggests that learned human motion distributions are a promising direction for egocentric camera and scene reconstruction.
Looking ahead, we plan to explore motion priors that, unlike diffusion models, allow efficient likelihood estimation during optimization, such as VAE-based formulations~\cite{rempe2021humor}.
Future work could also investigate feedforward models that jointly reconstruct scene and motion in a single pass, for which~\egomore~reconstructions could serve as pseudo-labels.
We believe that~\egomore~provides a strong foundation for future research on joint modeling of egocentric motion and 3D scenes.


\begin{ack}
This research was conducted at Stanford University with support from Panasonic Holdings Corporation and partial funding from NIH grant R01AG089169. The authors declare no competing interests beyond their listed affiliations.
\end{ack}

\clearpage

{
\bibliographystyle{abbrvnat}
\small
\bibliography{neurips_2026}
}

\appendix
\newpage
\section{Extended Evaluation and Results}
\label{sec:appendix_more_results}

Given that large-scale EgoExo4D and Nymeria datasets consist of egocentric sequences ranging from several minutes to over an hour, we extract 8-second segments for evaluation (following UniEgoMotion~\cite{uniegomotion}). To ensure a comprehensive analysis, our evaluation set contains two distinct subsets: (1) a representative subset sampled uniformly at random from the full dataset, and (2) a locomotion-heavy subset containing high camera translation and rotation. While the main paper reports results on the combined set containing 718 samples from EgoExo4D and 828 samples from Nymeria, we provide disaggregated results for each subset here. Additionally, we also evaluate a subset characterized by rotation-heavy camera movement which is challenging for visual SLAM methods~\cite{carlone2015initialization}. The disaggregated results are provided in Table ~\ref{tab:3d_recon_app} and ~\ref{tab:camera_pose_app}. We further evaluate per-frame depth accuracy of monocular and feedforward geometry models (Table~\ref{tab:depth_priors}), sensitivity to the choice of off-the-shelf priors (Table~\ref{tab:prior_sensitivity}), and performance on 3$\times$ longer sequences (Table~\ref{tab:long_seq}). Please refer to the supplementary video for additional qualitative results and comparisons with the baselines.

\section{Implementation Details}
We use RAFT~\cite{raft} to compute both forward and backward optical flow, incorporating both into the optimization objective. Long-range correspondences are obtained using CoTracker3~\cite{cotracker3} by seeding a $35 \times 35$ point grid at every 8-frame intervals and tracking them bi-directionally. Initial relative depth $\bs D^r$ and the metric scale prior $\bs S^m$ are provided by MoGe-2 models~\cite{moge2}. For the optimization objective, we set the loss weights to $1.0, 1.0, 5\times10^{-3}, 5\times10^{-3}$, and $5$ for $\mc{L}_\text{flow}$, $\mc{L}_\text{track}$, $\mc{L}_\text{scene}$, $\mc{L}_\text{depth}$, and $\mc{L}_\text{egocam}$, respectively. For the motion prior $\mc{L}_\text{motion}$, we sample the noise level $t$ in decreasing order following~\cite{lu2023dposer} and scale the loss weight according to the signal-to-noise ratio of the noised sample~\cite{sdsloss}. To handle outliers in $\mc{L}_\text{flow}$ and $\mc{L}_\text{track}$, we filter correspondences based on the Median Absolute Deviation (MAD) of the reprojection error. During each round of alternating optimization, we optimize $(\bs C, \bs S)$ for 1000 steps, followed by 100 steps of optimization for $\bs M$. Note that when $(\bs C, \bs S)$ is fixed, optimizing the motion for $T$ steps is equivalent to DDIM sampling of $\bs M$ for $T$ diffusion steps under the guidance of $\mc{L}_\text{egocam}$. We use the Adam optimizer with an initial learning rate of $0.01$, decaying to $0.001$ after 850 steps.

~\egomore~takes approximately 8\,min per 8-second clip on one NVIDIA L40S GPU: about 32\,s to compute the visuo-geometric priors, 3\,min and 4\,min for scene optimization in the first and second rounds, and 5\,s per round for motion optimization. This is slower than the feedforward baselines (a few seconds) and the optimization-based ViPE (about 1.5\,min). However, after one round of optimization, at less than half the total cost,~\egomore~already outperforms all baselines on metric 3D reconstruction on the EgoExo4D locomotion subset (Tabs.~\ref{tab:alternating_ablation} and~\ref{tab:3d_recon_app}).

Our motion prior $p(\bs M \vert \mc{A}(\bs C))$ is implemented as a conditional denoising diffusion model $f_\phi(\bs M_t, \mc{A}(\bs C))$. Following~\cite{uniegomotion}, we adopt an egocentric body motion representation that implicitly defines a canonical reference frame with a ground plane and gravity direction. The denoiser is implemented as a 1D UNet~\cite{gmd}, though we observed that transformer-based architecture~\cite{mdm} yields comparable performance. To evaluate the zero-shot motion generalization of~\egomore, we do not utilize motion annotations from EgoExo4D or Nymeria during training; and instead train exclusively on motion sequences from the AMASS dataset~\cite{amass}.

\begin{figure}[t]
\centering
\includegraphics[width=0.95\textwidth]{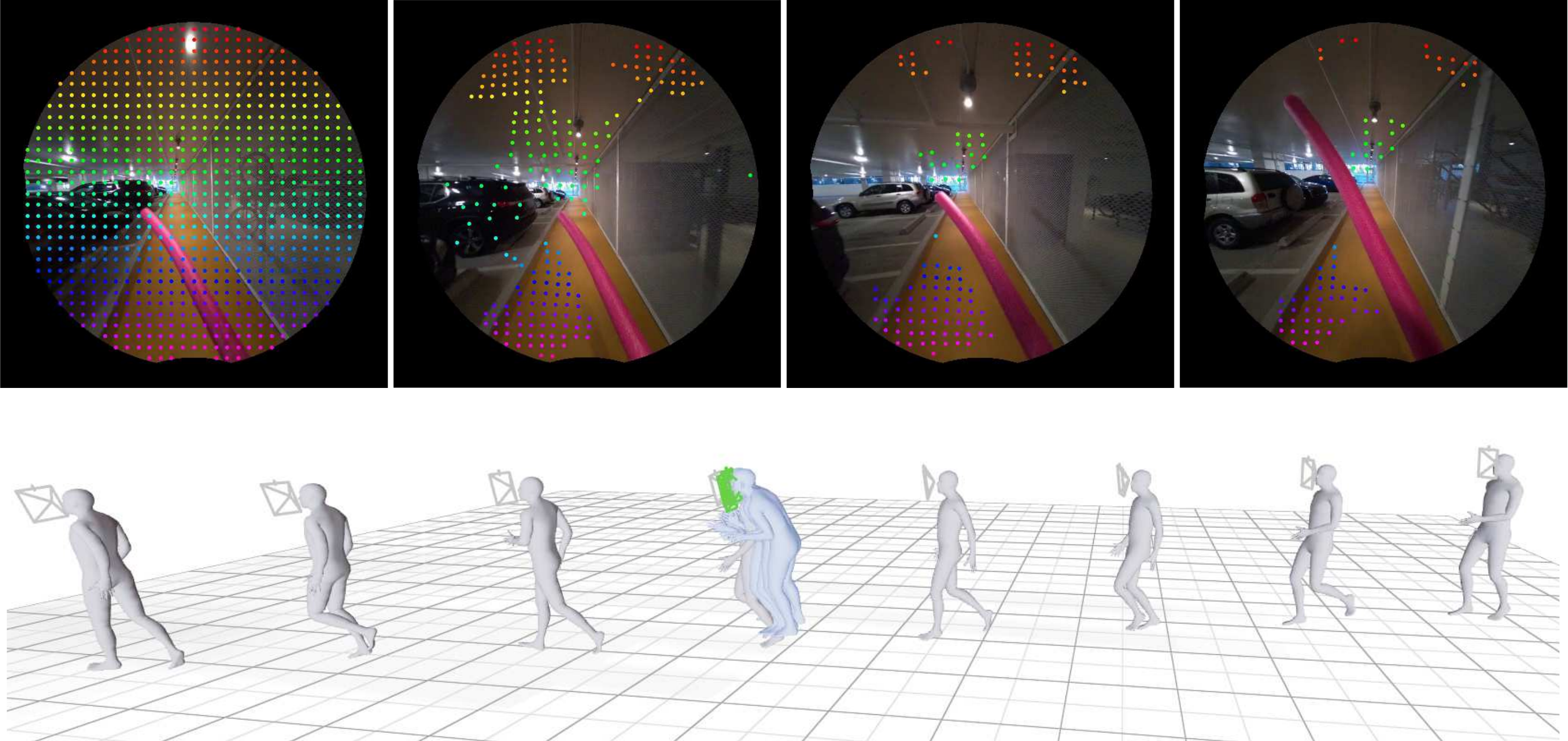}
\caption{\textbf{Point-tracking Failure}: Inaccurate long-range correspondences from the point-tracking model cause~\egomore~to predict stationary camera trajectory (green) and stationary motion (blue), whereas the groundtruth camera and motion trajectories (grey) move in a straight line.}
\label{fig:pointtracking_failure}
\end{figure}

\section{More Discussion on Motion Prior}
\label{sec:moremotionprior}
Our motion prior design (see Section 3.3) is motivated by the broader literature on inverse problems, guided sampling, and conditional generative models.
Recall that our objective is to reconstruct body motion $\bs M$, camera trajectory $\bs C$, and depth scales $\bs S$ that are consistent with the observations provided by the egocentric video $\bs I$.
Since the camera trajectory $\bs C$ is tightly coupled with the head trajectory $\mc{H}(\bs M)$ derived from the body motion (enforced via $\mc{L}_\text{egocam}$), we regularize the optimization of $\bs M$ (and consequently $\bs C$) using a prior on $\bs M$.
This formulation can be viewed as an inverse problem that aims to recover an unknown signal $x$ from noisy and incomplete observations $y = \mc{O}(x) + \epsilon$, where a data term $\mc{L}_\text{data}(\mc{O}(x), y)$ is minimized alongside a prior term $\mc{L}_\text{prior}(x)$.
In our context, we optimize $\bs M$ using data terms ($\mc{L}_\text{flow}$, $\mc{L}_\text{track}$, $\mc{L}_\text{egocam}$, etc.) regularized by the motion prior $p(\bs M \vert \cdot)$. Notably, the observations provide direct constraints only on a small subset of the motion $\bs M$ (i.e., the head trajectory), leaving the rest of the body pose highly dependent on the prior.

A straightforward approach to solving such inverse problems is to penalize the negative log-likelihood of $x$ during optimization ($\mc{L}_\text{prior}(x) = -\log p(x)$). Early works utilized simple pose priors modeled via Gaussian Mixture Models (GMMs) for single-frame tasks, such as fitting a 3D body model to 2D images~\cite{smplify,smplx}. While GMMs allow for exact likelihood calculation and efficient sampling, their modeling capacity is limited for high-dimensional, complex distributions like human motion. Recently, some works have used VAE-based~\cite{smplx,rempe2021humor} or diffusion-based~\cite{lu2023dposer} priors for their superior expressive power. However, since these models do not provide an explicit likelihood, incorporating them into a first-order optimization is non-trivial.
Consider a diffusion model that predicts $\mathbb{E}[x_0 \vert x_t]$, where the subscript denotes the noise level. One way to solve an inverse problem with such a model is to apply guidance~\cite{diffusionbeatsgans} during each denoising step:
\begin{align*}
\mathbb{E}[x_0 \vert x_t, y] = \mathbb{E}[x_0 \vert x_t] + \lambda \nabla_{x_t} \log p(y \vert x_t)
\end{align*}
where the second term provides guidance toward the observation $y$. ~\cite{diffusionbeatsgans} trained a separate classifier $p(y \vert x_t)$ for all noise levels to provide exact guidance. However, in many inverse problems, defining the guidance likelihood $p(y \vert x_t)$ on noisy data $x_t$ (rather than clean data $x_0$) is analytically intractable, as it requires marginalizing over $x_0$: $p(y \vert x_t) = \int p(y \vert x_0) p(x_0 \vert x_t) dx_0$.
While various approximations have been proposed~\cite{liu2023accelerating,diffusionpostsampling,peng2024improving,zirvi2024diffusion,mardani2023variational} that work well for simple pose or motion fitting with full-body 2D observations~\cite{rempe2021humor,smplify,smplx,lu2023dposer}, these approximations often fail when observations $y$ are weak and highly localized, such as controlling only a specific body part trajectory~\cite{mdmasprior,gmd,dno}, leading to suboptimal or physically implausible results.

Instead, recent works~\cite{mdmasprior,gmd} directly model the \textit{conditional} distribution $p(x \vert y)$. This not only circumvents the intractability of the guidance term but also simplifies the learning task, as conditional generation is generally easier to learn than unconditional generation~\cite{returnofuncond}. For egocentric reconstruction, prior methods~\cite{egoallo,uniegomotion,egoego,guzov2025hmd2} typically assume access to a gravity-aligned, metric camera trajectory $\bs C$ and model $p(\bs M \vert \bs C)$ using conditional diffusion. This approach is highly effective because the head trajectory strongly constrains the distribution of the remaining body joints, especially torso and lower body.

In our setting, however, we are also optimizing for $\bs C$. One possible solution is to use an unconditional motion prior $p(\bs M)$ and perform guided optimization using correspondence losses. But, as previously noted, this often leads to suboptimal results due to approximations. Instead, we use a conditional diffusion prior $p(\bs M \vert \mc{A}(\bs C))$, where the conditioning is based on the $\text{SE}(3)$-invariant trajectory $\mc{A}(\bs C)$. This allows us to incorporate the motion prior early in the optimization process. Following ~\cite{sdsloss}, we do not backpropagate gradients through the prior model itself to maintain optimization stability.

While this design provides effective regularization for $\bs C$ and improves 3D reconstruction (see Table ~\ref{tab:ablation_optim}, row 5), direct optimization often causes the motion to collapse to a solution that satisfies the trajectory $\bs C$ but disregards global physical plausibility, leading to worse motion metrics. We hypothesize this is due to simultaneous optimization, where gradients on $\bs M$ are often noisy relative to the data-term gradients on $\bs C$ during early optimization steps. While it is possible to mitigate this by carefully scheduling the loss weights as $\bs C$ approaches the optimal solution, such tuning is often brittle. Instead, we propose a simple alternating scheme that iterates between optimizing $(\bs C, \bs S)$ and $\bs M$. As shown in Table ~\ref{tab:ablation_optim} (row 6) in the paper and Table ~\ref{tab:alternating_ablation}, this strategy significantly improves both 3D reconstruction and motion reconstruction metrics.

\section{Limitations \& Future Work}

While~\egomore~achieves superior 3D scene reconstruction and accurate human motion estimation, it is not without limitations. We discuss several of them here and outline potential future directions.

The core design of~\egomore~relies on the robustness and generalization of low-level visuo-geometric models on challenging egocentric data. We observe that while optical flow generally provides stable short-range correspondences, long-range point tracking remains a primary source of failure. One mode of point-tracking failure arises in regions with monotonic textures or around handheld objects. These erroneous tracks can derail camera trajectory optimization, leading to poor local minima, as illustrated in Fig.~\ref{fig:pointtracking_failure}. A secondary source of failure involves frames with extreme motion blur, where both the relative depth and the metric scale prior $\bs S^m$ can become unreliable. This can result in sudden jitters or jumps in the estimated camera poses $\bs C$. Although these local inaccuracies typically do not compromise the global trajectory, they can degrade the qualitative consistency and temporal smoothness of the resulting 3D reconstruction and motion $\bs M$. Another minor failure mode is the lack of floor detection in \textit{all} frames. While this does not catastrophically degrade performance, it weakens the alignment between motion and 3D reconstruction, leaving the system to rely primarily on the egocentric camera trajectory.

Note that~\egomore~focuses on reconstructing scene and body pose from egocentric locomotion video input. This is motivated by the fact that torso and lower-body motion are highly correlated with head motion. However, wrist motion and hand manipulation are largely unrelated to the rest of the body motion. Some prior works in egocentric motion reconstruction, such as EgoAllo~\cite{egoallo} and HaMoS~\cite{hamos}, use off-the-shelf 3D hand predictors and guide motion generation with accurate hand trajectories to reconstruct whole-body motion. However, our focus is primarily on locomotion without egocentric trajectory input, where body dynamics are more strongly coupled. Future work could explore leveraging 3D scene context to reconstruct globally accurate hand motion alongside whole-body motion, which remains highly challenging without canonical trajectory.

Note that, like any SLAM method, dynamic objects violate our correspondence assumptions. We use Median Absolute Deviation (MAD) filtering to mitigate the effect of bad correspondences, which often arise from dynamic objects. However, when a large number of pixels in the input video belong to dynamic objects, MAD filtering may also remove valid correspondences. Many prior methods mask out dynamic objects using off-the-shelf estimators~\cite{vipe} or learned confidence masks~\cite{zhang2024monst3r,megasam,egomono4d}. While such dynamic-object masking did not provide significant benefits in our case, it can be easily incorporated into our pipeline by prompting the segmentation model used in our work with additional queries. Additionally,~\egomore~inference is relatively slow due to its optimization-based framework compared to feedforward geometry models. However, it provides more accurate reconstruction and can serve as a source of pseudo-labels to improve feedforward methods. Further improvements may also be achieved through explicit loop closure, akin to traditional SLAM pipelines.

Although the canonical camera trajectory $\bs C$ used in EgoAllo and UniEgoMotion contains more information than the $\text{SE}(3)$-invariant trajectory $\mc{A}(\bs C)$, our motion reconstruction still outperforms these baselines. We hypothesize that because canonical camera trajectory is highly informative of lower-body motion, any error in canonicalizing the predicted camera trajectory can significantly affect motion generation in the absence of a feedback loop. In contrast,~\egomore~conditions motion reconstruction on looser conditioning, thereby providing greater flexibility. However, removing canonicalization also discards crucial information about body motion, forcing the model to rely more on guidance from $\mc{L}_\text{scene}$ and $\mc{L}_\text{egocam}$ (both in the canonical frame), which is often brittle (see Sec.~\ref{sec:moremotionprior}). Future work could explore more effective ways to incorporate canonicalization information into motion generation.

\begin{table*}[t]\centering\footnotesize
\setlength{\tabcolsep}{4.0pt}
\ra{1.1} 
\caption{\textbf{3D Reconstruction}: Comparison across metric and scale-invariant 3D reconstruction on three subsets (random, locomotion-heavy, rotation-heavy) of each dataset. Lower values ($\downarrow$) indicate better performance.}
\label{tab:3d_recon_app}
\begin{tabular}{lllcccccccc} 
\toprule
& & & \multicolumn{4}{c}{Metric Scale} & \multicolumn{4}{c}{Scale Invariant} \\
\cmidrule(lr){4-7} \cmidrule(lr){8-11}
& & & \multicolumn{2}{c}{Acc} & \multicolumn{2}{c}{Comp} & \multicolumn{2}{c}{Acc} & \multicolumn{2}{c}{Comp} \\
& & Method & Mean & Median & Mean & Median & Mean & Median & Mean & Median \\
\midrule
\multirow{15}{*}{\rotatebox[origin=c]{90}{\textbf{EgoExo4D}}} 
& \multirow{5}{*}{\rotatebox[origin=c]{90}{Random}} 
& VGGT~\cite{vggt} & $-$ & $-$ & $-$ & $-$ & $0.432$ & $0.382$ & $0.425$ & $0.231$ \\
& & CUT3R~\cite{cut3r} & $0.402$ & $0.372$ & $0.613$ & $0.480$ & $0.385$ & $0.332$ & $0.314$ & $0.214$ \\
& & ViPE~\cite{vipe} & $0.682$ & $0.545$ & $0.829$ & $0.607$ & $0.384$ & $0.329$ & $0.320$ & $0.173$ \\
& & EgoMono4D~\cite{egomono4d} & $0.638$ & $0.565$ & $0.637$ & $0.510$ & $0.431$ & $0.390$ & $0.533$ & $0.384$ \\
& & Ours & $\mathbf{0.316}$ & $\mathbf{0.254}$ & $\mathbf{0.163}$ & $\mathbf{0.110}$ & $\mathbf{0.267}$ & $\mathbf{0.211}$ & $\mathbf{0.135}$ & $\mathbf{0.081}$ \\
\cmidrule(lr){2-11}
& \multirow{5}{*}{\rotatebox[origin=c]{90}{Locomotion}} 
& VGGT~\cite{vggt} & $-$ & $-$ & $-$ & $-$ & $0.507$ & $0.453$ & $0.454$ & $0.220$ \\
& & CUT3R~\cite{cut3r} & $0.463$ & $0.421$ & $0.713$ & $0.531$ & $0.433$ & $0.367$ & $0.367$ & $0.237$ \\
& & ViPE~\cite{vipe} & $1.120$ & $0.765$ & $0.808$ & $0.540$ & $0.464$ & $0.402$ & $0.382$ & $0.214$ \\
& & EgoMono4D~\cite{egomono4d} & $0.599$ & $0.524$ & $0.773$ & $0.580$ & $0.506$ & $0.455$ & $0.651$ & $0.440$ \\
& & Ours & $\mathbf{0.396}$ & $\mathbf{0.320}$ & $\mathbf{0.209}$ & $\mathbf{0.137}$ & $\mathbf{0.349}$ & $\mathbf{0.278}$ & $\mathbf{0.180}$ & $\mathbf{0.115}$ \\
\cmidrule(lr){2-11}
& \multirow{5}{*}{\rotatebox[origin=c]{90}{Rotation}} 
& VGGT~\cite{vggt} & $-$ & $-$ & $-$ & $-$ & $0.667$ & $0.594$ & $0.604$ & $0.330$ \\
& & CUT3R~\cite{cut3r} & $0.680$ & $0.635$ & $1.036$ & $0.755$ & $0.590$ & $0.518$ & $0.585$ & $0.375$ \\
& & ViPE~\cite{vipe} & $0.708$ & $0.648$ & $0.922$ & $0.628$ & $0.626$ & $0.557$ & $0.608$ & $0.388$ \\
& & EgoMono4D~\cite{egomono4d} & $0.790$ & $0.738$ & $1.224$ & $0.916$ & $0.705$ & $0.653$ & $1.015$ & $0.687$ \\
& & Ours & $\mathbf{0.638}$ & $\mathbf{0.552}$ & $\mathbf{0.377}$ & $\mathbf{0.261}$ & $\mathbf{0.566}$ & $\mathbf{0.485}$ & $\mathbf{0.359}$ & $\mathbf{0.247}$ \\
\midrule 
\multirow{15}{*}{\rotatebox[origin=c]{90}{\textbf{Nymeria}}} 
& \multirow{5}{*}{\rotatebox[origin=c]{90}{Random}} 
& VGGT~\cite{vggt} & $-$ & $-$ & $-$ & $-$ & $0.320$ & $0.276$ & $0.303$ & $0.155$ \\
& & CUT3R~\cite{cut3r} & $0.337$ & $0.289$ & $0.369$ & $0.242$ & $0.321$ & $0.270$ & $0.293$ & $0.183$ \\
& & ViPE~\cite{vipe} & $0.332$ & $0.298$ & $0.492$ & $0.307$ & $0.319$ & $0.274$ & $0.250$ & $0.128$ \\
& & EgoMono4D~\cite{egomono4d} & $0.507$ & $0.427$ & $0.534$ & $0.379$ & $0.369$ & $0.327$ & $0.516$ & $0.357$ \\
& & Ours & $\mathbf{0.227}$ & $\mathbf{0.174}$ & $\mathbf{0.117}$ & $\mathbf{0.067}$ & $\mathbf{0.201}$ & $\mathbf{0.153}$ & $\mathbf{0.101}$ & $\mathbf{0.055}$ \\
\cmidrule(lr){2-11}
& \multirow{5}{*}{\rotatebox[origin=c]{90}{Locomotion}} 
& VGGT~\cite{vggt} & $-$ & $-$ & $-$ & $-$ & $0.354$ & $0.305$ & $0.318$ & $0.148$ \\
& & CUT3R~\cite{cut3r} & $0.357$ & $0.309$ & $0.356$ & $0.217$ & $0.357$ & $0.303$ & $0.296$ & $0.181$ \\
& & ViPE~\cite{vipe} & $3.118$ & $2.712$ & $1.028$ & $0.820$ & $0.373$ & $0.323$ & $0.296$ & $0.155$ \\
& & EgoMono4D~\cite{egomono4d} & $0.476$ & $0.410$ & $0.587$ & $0.408$ & $0.423$ & $0.386$ & $0.578$ & $0.397$ \\
& & Ours & $\mathbf{0.269}$ & $\mathbf{0.206}$ & $\mathbf{0.146}$ & $\mathbf{0.079}$ & $\mathbf{0.248}$ & $\mathbf{0.191}$ & $\mathbf{0.135}$ & $\mathbf{0.075}$ \\
\cmidrule(lr){2-11}
& \multirow{5}{*}{\rotatebox[origin=c]{90}{Rotation}} 
& VGGT~\cite{vggt} & $-$ & $-$ & $-$ & $-$ & $0.383$ & $0.338$ & $0.331$ & $0.164$ \\
& & CUT3R~\cite{cut3r} & $0.336$ & $0.287$ & $0.316$ & $0.193$ & $0.341$ & $0.287$ & $0.261$ & $0.158$ \\
& & ViPE~\cite{vipe} & $0.386$ & $0.332$ & $0.474$ & $0.269$ & $0.371$ & $0.325$ & $0.375$ & $0.227$ \\
& & EgoMono4D~\cite{egomono4d} & $0.449$ & $0.388$ & $0.516$ & $0.335$ & $0.408$ & $0.372$ & $0.524$ & $0.334$ \\
& & Ours & $\mathbf{0.305}$ & $\mathbf{0.236}$ & $\mathbf{0.156}$ & $\mathbf{0.088}$ & $\mathbf{0.288}$ & $\mathbf{0.227}$ & $\mathbf{0.164}$ & $\mathbf{0.097}$ \\
\bottomrule
\end{tabular}
\vspace{-4mm}
\end{table*}

\begin{table*}[t]\centering\footnotesize
\setlength{\tabcolsep}{10.0pt}
\ra{1.1} 
\caption{\textbf{Camera Pose Estimation}: Comparison across metric and scale-invariant camera pose estimation on three subsets (random, locomotion-heavy, rotation-heavy) of each dataset. Lower values ($\downarrow$) indicate better performance.}
\label{tab:camera_pose_app}
\begin{tabular}{lllcccccc} 
\toprule
& & & \multicolumn{3}{c}{Metric Scale} & \multicolumn{3}{c}{Scale Invariant} \\
\cmidrule(lr){4-6} \cmidrule(lr){7-9}
& & Method & ATE & RPE Rot & RPE Trans & ATE & RPE Rot & RPE Trans \\
\midrule
\multirow{18}{*}{\rotatebox[origin=c]{90}{\textbf{EgoExo4D}}} 
& \multirow{6}{*}{\rotatebox[origin=c]{90}{Random}} 
& EgoEgo~\cite{egoego} & $0.246$ & $2.652$ & $0.051$ & $0.171$ & $2.652$ & $0.053$ \\
& & VGGT~\cite{vggt} & $-$ & $-$ & $-$ & $0.156$ & $2.510$ & $0.049$ \\
& & CUT3R~\cite{cut3r} & $0.269$ & $2.341$ & $0.075$ & $0.174$ & $2.341$ & $0.045$ \\
& & ViPE~\cite{vipe} & $0.308$ & $1.378$ & $0.082$ & $\mathbf{0.053}$ & $1.378$ & $\mathbf{0.020}$ \\
& & EgoMono4D~\cite{egomono4d} & $0.971$ & $2.829$ & $0.252$ & $0.204$ & $2.829$ & $0.049$ \\
& & Ours & $\mathbf{0.131}$ & $\mathbf{1.240}$ & $\mathbf{0.034}$ & $0.082$ & $\mathbf{1.240}$ & $0.024$ \\
\cmidrule(lr){2-9}
& \multirow{6}{*}{\rotatebox[origin=c]{90}{Locomotion}} 
& EgoEgo~\cite{egoego} & $0.589$ & $7.876$ & $0.126$ & $0.512$ & $7.876$ & $0.111$ \\
& & VGGT~\cite{vggt} & $-$ & $-$ & $-$ & $0.343$ & $4.930$ & $0.106$ \\
& & CUT3R~\cite{cut3r} & $0.556$ & $5.796$ & $0.160$ & $0.410$ & $5.796$ & $0.105$ \\
& & ViPE~\cite{vipe} & $0.540$ & $\mathbf{3.711}$ & $0.190$ & $\mathbf{0.169}$ & $\mathbf{3.711}$ & $\mathbf{0.044}$ \\
& & EgoMono4D~\cite{egomono4d} & $1.717$ & $6.002$ & $0.439$ & $0.481$ & $6.002$ & $0.103$ \\
& & Ours & $\mathbf{0.318}$ & $3.988$ & $\mathbf{0.095}$ & $0.237$ & $3.988$ & $0.070$ \\
\cmidrule(lr){2-9}
& \multirow{6}{*}{\rotatebox[origin=c]{90}{Rotation}} 
& EgoEgo~\cite{egoego} & $1.015$ & $14.620$ & $0.219$ & $0.831$ & $14.620$ & $0.159$ \\
& & VGGT~\cite{vggt} & $-$ & $-$ & $-$ & $0.714$ & $13.338$ & $0.204$ \\
& & CUT3R~\cite{cut3r} & $1.144$ & $14.969$ & $0.380$ & $0.737$ & $14.969$ & $0.192$ \\
& & ViPE~\cite{vipe} & $1.416$ & $10.668$ & $0.580$ & $\mathbf{0.404}$ & $10.668$ & $\mathbf{0.106}$ \\
& & EgoMono4D~\cite{egomono4d} & $2.286$ & $10.967$ & $0.620$ & $0.757$ & $10.967$ & $0.164$ \\
& & Ours & $\mathbf{0.579}$ & $\mathbf{10.199}$ & $\mathbf{0.209}$ & $0.417$ & $\mathbf{10.199}$ & $0.140$ \\
\midrule
\multirow{18}{*}{\rotatebox[origin=c]{90}{\textbf{Nymeria}}} 
& \multirow{6}{*}{\rotatebox[origin=c]{90}{Random}} 
& EgoEgo~\cite{egoego} & $0.352$ & $3.057$ & $0.053$ & $0.261$ & $3.057$ & $0.073$ \\
& & VGGT~\cite{vggt} & $-$ & $-$ & $-$ & $0.157$ & $1.893$ & $0.056$ \\
& & CUT3R~\cite{cut3r} & $0.323$ & $1.831$ & $0.071$ & $0.217$ & $1.831$ & $0.051$ \\
& & ViPE~\cite{vipe} & $\mathbf{0.173}$ & $1.235$ & $0.042$ & $\mathbf{0.062}$ & $1.235$ & $\mathbf{0.020}$ \\
& & EgoMono4D~\cite{egomono4d} & $0.966$ & $2.667$ & $0.238$ & $0.254$ & $2.667$ & $0.065$ \\
& & Ours & $0.190$ & $\mathbf{1.104}$ & $\mathbf{0.032}$ & $0.127$ & $\mathbf{1.104}$ & $0.032$ \\
\cmidrule(lr){2-9}
& \multirow{6}{*}{\rotatebox[origin=c]{90}{Locomotion}} 
& EgoEgo~\cite{egoego} & $0.901$ & $6.774$ & $0.138$ & $0.772$ & $6.774$ & $0.208$ \\
& & VGGT~\cite{vggt} & $-$ & $-$ & $-$ & $0.393$ & $4.043$ & $0.159$ \\
& & CUT3R~\cite{cut3r} & $0.621$ & $3.788$ & $0.167$ & $0.465$ & $3.788$ & $0.136$ \\
& & ViPE~\cite{vipe} & $\mathbf{0.388}$ & $\mathbf{3.202}$ & $0.133$ & $\mathbf{0.185}$ & $\mathbf{3.202}$ & $\mathbf{0.060}$ \\
& & EgoMono4D~\cite{egomono4d} & $1.666$ & $5.131$ & $0.408$ & $0.618$ & $5.131$ & $0.164$ \\
& & Ours & $0.463$ & $3.208$ & $\mathbf{0.090}$ & $0.352$ & $3.208$ & $0.098$ \\
\cmidrule(lr){2-9}
& \multirow{6}{*}{\rotatebox[origin=c]{90}{Rotation}} 
& EgoEgo~\cite{egoego} & $0.612$ & $9.803$ & $0.108$ & $0.554$ & $9.803$ & $0.131$ \\
& & VGGT~\cite{vggt} & $-$ & $-$ & $-$ & $0.347$ & $\mathbf{5.445}$ & $0.153$ \\
& & CUT3R~\cite{cut3r} & $0.550$ & $6.132$ & $0.211$ & $0.396$ & $6.132$ & $0.134$ \\
& & ViPE~\cite{vipe} & $0.635$ & $6.397$ & $0.272$ & $\mathbf{0.248}$ & $6.397$ & $\mathbf{0.078}$ \\
& & EgoMono4D~\cite{egomono4d} & $1.738$ & $7.816$ & $0.526$ & $0.516$ & $7.816$ & $0.144$ \\
& & Ours & $\mathbf{0.467}$ & $\mathbf{6.056}$ & $\mathbf{0.122}$ & $0.404$ & $6.056$ & $0.115$ \\
\bottomrule
\end{tabular}
\vspace{-4mm}
\end{table*}

\clearpage

\begin{table*}[t]\centering\footnotesize
\setlength{\tabcolsep}{4.0pt}
\ra{1.1}
\caption{\textbf{Per-Frame Depth Accuracy}: Depth predicted by multi-view feedforward geometry models (top) and monocular depth models (bottom) on all frames of the EgoExo4D locomotion subset, using the full rectified frames ($512\times512$) and their central $384\times384$ crop. Monocular depth models are more accurate than feedforward geometry models on every metric in both settings, which motivates their use as our geometry basis.}
\label{tab:depth_priors}
\begin{tabular}{lcccccccc}
\toprule
& \multicolumn{4}{c}{Full frame} & \multicolumn{4}{c}{Center crop} \\
\cmidrule(lr){2-5} \cmidrule(lr){6-9}
& \multicolumn{2}{c}{Metric} & \multicolumn{2}{c}{Scale Invariant} & \multicolumn{2}{c}{Metric} & \multicolumn{2}{c}{Scale Invariant} \\
\cmidrule(lr){2-3} \cmidrule(lr){4-5} \cmidrule(lr){6-7} \cmidrule(lr){8-9}
Method & AbsRel$\downarrow$ & $\delta_{1.25}\uparrow$ & AbsRel$\downarrow$ & $\delta_{1.25}\uparrow$ & AbsRel$\downarrow$ & $\delta_{1.25}\uparrow$ & AbsRel$\downarrow$ & $\delta_{1.25}\uparrow$ \\
\midrule
VGGT~\cite{vggt} & $-$ & $-$ & $0.543$ & $0.503$ & $-$ & $-$ & $0.353$ & $0.576$ \\
CUT3R~\cite{cut3r} & $0.456$ & $0.372$ & $0.175$ & $0.772$ & $0.445$ & $0.418$ & $0.292$ & $0.658$ \\
\midrule
MoGe-2-L~\cite{moge2} ({\scriptsize Used in~\egomore}) & $0.313$ & $0.551$ & $0.158$ & $0.809$ & $0.251$ & ${0.651}$ & ${0.130}$ & ${0.853}$ \\
UniDepth-v2-L~\cite{piccinelli2025unidepthv2} ({\scriptsize Used in ViPE}) & $0.262$ & ${0.656}$ & $0.153$ & $0.824$ & $0.285$ & $0.619$ & $0.138$ & $0.843$ \\
Metric3D-v2-L~\cite{hu2024metric3d} & ${0.228}$ & $0.632$ & ${0.146}$ & ${0.828}$ & ${0.226}$ & $0.618$ & $0.139$ & $0.841$ \\
\bottomrule
\end{tabular}
\vspace{-4mm}
\end{table*}

\begin{table*}[t]\centering\footnotesize
\setlength{\tabcolsep}{4.0pt}
\ra{1.1}
\caption{\textbf{Sensitivity to Off-the-Shelf Priors}: We replace the point tracker or the depth model in the scene-only setting (row 3 of Tab.~\ref{tab:ablation_optim}). Lower values ($\downarrow$) indicate better performance. Weaker priors degrade the results gradually, without catastrophic failure. The weaker tracker has the larger effect, mainly on ATE and RPE Trans, since less reliable long-range correspondences lead to more drift.}
\label{tab:prior_sensitivity}
\begin{tabular}{lcccccc}
\toprule
 & \multirow{2}{*}{ATE} & \multirow{2}{*}{RPE Trans} & \multicolumn{2}{c}{Acc} & \multicolumn{2}{c}{Comp} \\
\cmidrule(lr){4-5} \cmidrule(lr){6-7}
Tracker + Depth & & & Mean & Median & Mean & Median \\
\midrule
CoTracker3 + MoGe-2 (ours) & $\mathbf{0.353}$ & $\mathbf{0.103}$ & $\mathbf{0.414}$ & $\mathbf{0.337}$ & $\mathbf{0.210}$ & $\mathbf{0.135}$ \\
CoTracker3 + MoGe & $0.389$ & $0.123$ & $0.415$ & $0.345$ & $0.213$ & $0.141$ \\
CoTracker + MoGe-2 & $0.505$ & $0.171$ & $0.454$ & $0.376$ & $0.243$ & $0.147$ \\
\bottomrule
\end{tabular}
\vspace{-4mm}
\end{table*}

\begin{table*}[t]\centering\footnotesize
\setlength{\tabcolsep}{4.0pt}
\ra{1.1}
\caption{\textbf{Long Sequences}: Results on 24\,s EgoExo4D locomotion sequences, $3\times$ the length of the main evaluation clips, with metric scale alignment. Lower values ($\downarrow$) indicate better performance. Our motion prior, trained on 8\,s windows, is applied over overlapping windows with MultiDiffusion~\cite{bar2023multidiffusion}; the rest of the optimization is unchanged. VGGT runs out of memory on a 48\,GB NVIDIA L40S GPU. Even without explicit loop closure, \egomore~remains best on all reconstruction metrics and on ATE, which we attribute to the long-range tracks from \textit{offline} CoTracker3~\cite{cotracker3}.}
\label{tab:long_seq}
\begin{tabular}{lccccccc}
\toprule
& \multicolumn{2}{c}{Acc} & \multicolumn{2}{c}{Comp} & \multirow{2}{*}{ATE} & \multirow{2}{*}{RPE Rot} & \multirow{2}{*}{RPE Trans} \\
\cmidrule(lr){2-3} \cmidrule(lr){4-5}
Method & Mean & Median & Mean & Median & & & \\
\midrule
VGGT~\cite{vggt} & \multicolumn{7}{c}{out of memory} \\
CUT3R~\cite{cut3r} & $0.614$ & $0.506$ & $0.621$ & $0.401$ & $1.226$ & $10.879$ & $\mathbf{0.251}$ \\
ViPE~\cite{vipe} & $0.735$ & $0.633$ & $0.732$ & $0.474$ & $1.668$ & $\mathbf{8.308}$ & $0.416$ \\
EgoMono4D~\cite{egomono4d} & $0.786$ & $0.699$ & $0.836$ & $0.554$ & $2.411$ & $8.851$ & $0.415$ \\
Ours & $\mathbf{0.505}$ & $\mathbf{0.398}$ & $\mathbf{0.212}$ & $\mathbf{0.105}$ & $\mathbf{0.913}$ & $9.605$ & $0.343$ \\
\bottomrule
\end{tabular}
\vspace{-4mm}
\end{table*}

\newpage
\clearpage
\section*{NeurIPS Paper Checklist}

\begin{enumerate}

\item {\bf Claims}
    \item[] Question: Do the main claims made in the abstract and introduction accurately reflect the paper's contributions and scope?
    \item[] Answer: \answerYes{} 
    \item[] Justification: The abstract and introduction clearly state the contributions in joint optimization of camera poses, scene geometry, and human motion from egocentric locomotion video, supported by experimental results on EgoExo4D and Nymeria sequences (Tables 1-3) and ablation studies (Table 4).

\item {\bf Limitations}
    \item[] Question: Does the paper discuss the limitations of the work performed by the authors?
    \item[] Answer: \answerYes{} 
    \item[] Justification: The Appendix discusses the main limitations of ~\egomore, including its reliance on visuo-geometric priors, long optimization time, and lack of hand motion optimization.

\item {\bf Theory assumptions and proofs}
    \item[] Question: For each theoretical result, does the paper provide the full set of assumptions and a complete (and correct) proof?
    \item[] Answer: \answerNA{} 
    \item[] Justification: NA.

    \item {\bf Experimental result reproducibility}
    \item[] Question: Does the paper fully disclose all the information needed to reproduce the main experimental results of the paper to the extent that it affects the main claims and/or conclusions of the paper (regardless of whether the code and data are provided or not)?
    \item[] Answer: \answerYes{} 
    \item[] Justification: The paper provides detailed implementation details in the Appendix, including the specific off-the-shelf models used, loss weights, optimizer settings, learning rate schedule, and the number of optimization steps per alternating round. The training data (AMASS) and evaluation datasets (EgoExo4D, Nymeria) are publicly available, and we commit to releasing code.

\item {\bf Open access to data and code}
    \item[] Question: Does the paper provide open access to the data and code, with sufficient instructions to faithfully reproduce the main experimental results, as described in supplemental material?
    \item[] Answer: \answerYes{} 
    \item[] Justification: The evaluation datasets (EgoExo4D, Nymeria) and the training dataset (AMASS) used for the motion prior are publicly available. We mention in the Appendix that the code will be released.

\item {\bf Experimental setting/details}
    \item[] Question: Does the paper specify all the training and test details (e.g., data splits, hyperparameters, how they were chosen, type of optimizer) necessary to understand the results?
    \item[] Answer: \answerYes{} 
    \item[] Justification: The Appendix provides comprehensive implementation details, including the loss weights, learning rate schedule, etc. The data splits and evaluation protocol are also described, and will be released with our code for reproducibility.

\item {\bf Experiment statistical significance}
    \item[] Question: Does the paper report error bars suitably and correctly defined or other appropriate information about the statistical significance of the experiments?
    \item[] Answer: \answerNA{} 
    \item[] Justification: Due to the computational cost of evaluation, prior works in this area do not report error bars, and we follow this convention. Instead, we report both mean and median metrics for 3D reconstruction (following prior works) and provide disaggregated results across three subsets (random, locomotion-heavy, rotation-heavy) in the Appendix to characterize performance variation.

\item {\bf Experiments compute resources}
    \item[] Question: For each experiment, does the paper provide sufficient information on the computer resources (type of compute workers, memory, time of execution) needed to reproduce the experiments?
    \item[] Answer: \answerYes{} 
    \item[] Justification: The Appendix reports that two rounds of alternating optimization take approximately 8 minutes per sample on NVIDIA L40S 48G GPU, including the one-time computation of visuo-geometric priors. 
    
\item {\bf Code of ethics}
    \item[] Question: Does the research conducted in the paper conform, in every respect, with the NeurIPS Code of Ethics \url{https://neurips.cc/public/EthicsGuidelines}?
    \item[] Answer: \answerYes{} 
    \item[] Justification:  The research conforms with the NeurIPS Code of Ethics. We use only publicly available datasets (EgoExo4D, Nymeria, AMASS) collected with appropriate consent by their original authors, and the work poses no foreseeable ethical concerns.

\item {\bf Broader impacts}
    \item[] Question: Does the paper discuss both potential positive societal impacts and negative societal impacts of the work performed?
    \item[] Answer: \answerYes{} 
    \item[] Justification: The paper's introduction highlights positive societal impacts in AR/VR, assistive technologies, healthcare, and robot learning. As with any egocentric perception system, potential negative impacts include privacy concerns associated with capture and reconstruction of personal environments; these concerns are widely explored in the egocentric vision community, and our work does not introduce any new concerns beyond those already discussed.
    
\item {\bf Safeguards}
    \item[] Question: Does the paper describe safeguards that have been put in place for responsible release of data or models that have a high risk for misuse (e.g., pre-trained language models, image generators, or scraped datasets)?
    \item[] Answer: \answerNA{} 
    \item[] Justification: The paper does not release any new datasets or pre-trained models with a high risk for misuse. The proposed method is an optimization framework that builds on existing publicly available models and datasets.

\item {\bf Licenses for existing assets}
    \item[] Question: Are the creators or original owners of assets (e.g., code, data, models), used in the paper, properly credited and are the license and terms of use explicitly mentioned and properly respected?
    \item[] Answer: \answerYes{} 
    \item[] Justification: All existing assets used in the paper, including datasets (EgoExo4D, Nymeria, AMASS) and pre-trained models (RAFT, CoTracker3, MoGe-2, SAM 2), are properly cited with references to their original publications. We use these assets in accordance with their respective licenses and terms of use.

\item {\bf New assets}
    \item[] Question: Are new assets introduced in the paper well documented and is the documentation provided alongside the assets?
    \item[] Answer: \answerNA{} 
    \item[] Justification: The paper does not release any new datasets or models at submission time. Code will be released with appropriate documentation.

\item {\bf Crowdsourcing and research with human subjects}
    \item[] Question: For crowdsourcing experiments and research with human subjects, does the paper include the full text of instructions given to participants and screenshots, if applicable, as well as details about compensation (if any)? 
    \item[] Answer: \answerNA{} 
    \item[] Justification: The paper does not involve crowdsourcing or research with human subjects. All experiments are conducted on existing publicly available datasets.

\item {\bf Institutional review board (IRB) approvals or equivalent for research with human subjects}
    \item[] Question: Does the paper describe potential risks incurred by study participants, whether such risks were disclosed to the subjects, and whether Institutional Review Board (IRB) approvals (or an equivalent approval/review based on the requirements of your country or institution) were obtained?
    \item[] Answer: \answerNA{} 
    \item[] Justification: The paper does not involve research with human subjects. All experiments use existing publicly available datasets that were collected by their original authors with appropriate approvals.

\item {\bf Declaration of LLM usage}
    \item[] Question: Does the paper describe the usage of LLMs if it is an important, original, or non-standard component of the core methods in this research? Note that if the LLM is used only for writing, editing, or formatting purposes and does \emph{not} impact the core methodology, scientific rigor, or originality of the research, declaration is not required.
    \item[] Answer: \answerNA{} 
    \item[] Justification: LLMs are not used as any important, original, or non-standard component of the core methodology in this research.

\end{enumerate}

\end{document}